\documentclass[acmsmall]{acmart}

\usepackage{rotating}
\usepackage{multirow,graphicx}
\usepackage{float}
\usepackage{subfigure}
\usepackage{amsmath}
\usepackage[normalem]{ulem}
\newcommand\hl{%
  \bgroup
  \expandafter\def\csname sout\space\endcsname{\bgroup \ULdepth =-.8ex \ULset}%
  \markoverwith{\textcolor{yellow}{\rule[-.5ex]{.1pt}{2.5ex}}}%
  \ULon}
\AtBeginDocument{%
  }

\setcopyright{acmcopyright}
\copyrightyear{2023}
\acmYear{2023}
\acmDOI{XXXXXXX.XXXXXXX}

\begin{document}

\title{GANonymization: A GAN-based Face Anonymization Framework for Preserving Emotional Expressions}

\author{Fabio Hellmann}
\email{fabio.hellmann@informatik.uni-augsburg.de}
\orcid{0000-0001-6404-0827}
\affiliation{%
  \institution{University of Augsburg}
  \streetaddress{Universitaetsstrasse 6a}
  \city{Augsburg}
  \state{Bavaria}
  \country{Germany}
  \postcode{86159}
}
\author{Silvan Mertes}
\email{silvan.mertes@informatik.uni-augsburg.de}
\orcid{0000-0001-5230-5218}
\affiliation{%
  \institution{University of Augsburg}
  \streetaddress{Universitaetsstrasse 6a}
  \city{Augsburg}
  \state{Bavaria}
  \country{Germany}
  \postcode{86159}
}
\author{Mohamed Benouis}
\email{mohamed.benouis@informatik.uni-augsburg.de}
\orcid{0000-0002-9107-9329}
\affiliation{%
  \institution{University of Augsburg}
  \streetaddress{Universitaetsstrasse 6a}
  \city{Augsburg}
  \state{Bavaria}
  \country{Germany}
  \postcode{86159}
}
\author{Alexander Hustinx}
\email{ahustinx@uni-bonn.de}
\orcid{0000-0003-4592-3979}
\affiliation{%
  \institution{University of Bonn}
  \streetaddress{Venusberg-Campus 1}
  \city{Bonn}
  \state{North Rhine-Westphalia}
  \country{Germany}
  \postcode{53127}
}
\author{Tzung-Chien Hsieh}
\email{thsieh@uni-bonn.de}
\orcid{0000-0003-3828-4419}
\affiliation{%
  \institution{University of Bonn}
  \streetaddress{Venusberg-Campus 1}
  \city{Bonn}
  \state{North Rhine-Westphalia}
  \country{Germany}
  \postcode{53127}
}
\author{Cristina Conati}
\email{conati@cs.ubc.ca}
\orcid{0000-0002-8434-9335}
\affiliation{%
  \institution{University of British Columbia}
  \streetaddress{2366 Main Mall Vancouver}
  \city{Vancouver}
  \state{BC}
  \country{Canada}
  \postcode{V6T1Z4}
}
\author{Peter Krawitz}
\email{pkrawitz@uni-bonn.de}
\orcid{0000-0002-3194-8625}
\affiliation{%
  \institution{University of Bonn}
  \streetaddress{Venusberg-Campus 1}
  \city{Bonn}
  \state{North Rhine-Westphalia}
  \country{Germany}
  \postcode{53127}
}
\author{Elisabeth André}
\email{andre@informatik.uni-augsburg.de}
\orcid{0000-0002-2367-162X}
\affiliation{%
  \institution{University of Augsburg}
  \streetaddress{Universitaetsstrasse 6a}
  \city{Augsburg}
  \state{Bavaria}
  \country{Germany}
  \postcode{86159}
}

\renewcommand{\shortauthors}{Hellmann et al.}

\begin{abstract}
In recent years, the increasing availability of personal data has raised concerns regarding privacy and security. 
One of the critical processes to address these concerns is data anonymization, which aims to protect individual privacy and prevent the release of sensitive information. 
This research focuses on the importance of face anonymization. 
Therefore, we introduce GANonymization, a novel face anonymization framework with facial expression-preserving abilities.
Our approach is based on a high-level representation of a face, which is synthesized into an anonymized version based on a generative adversarial network (GAN).
The effectiveness of the approach was assessed by evaluating its performance in removing identifiable facial attributes to increase the anonymity of the given individual face.
Additionally, the performance of preserving facial expressions was evaluated on several affect recognition datasets and outperformed the state-of-the-art methods in most categories.
Finally, our approach was analyzed for its ability to remove various facial traits, such as jewelry, hair color, and multiple others. Here, it demonstrated reliable performance in removing these attributes.
Our results suggest that GANonymization is a promising approach for anonymizing faces while preserving facial expressions.
\end{abstract}



\received{5 April 2023}

\maketitle

\section{Introduction}
In the current machine learning landscape, models are getting more and more complex.
This complexity places a significant demand on the availability of large, high-quality datasets, particularly when leveraging deep learning (DL) techniques.
However, building such datasets is not always easy - besides the time-consuming process of acquiring and annotating data, privacy is a serious obstacle here. 
While extensive datasets exist for non-sensitive data, the acquisition of data for sensitive use cases, especially those involving human data, is an intricate task when the subjects' privacy needs to be ensured.
Particularly when it comes to scenarios involving the human face, it is generally a hard task to collect appropriate data, especially if datasets are to be made publicly available.
On the other hand, developing DL models that use images of human faces offers promising opportunities.
For instance, assessing affective states like emotions or stress might be beneficial to infer more serious conditions, such as chronic overload or depression, and react accordingly.
However, not only training data for DL algorithms run the risk of violating humans' privacy - it is inference data too. 
When employing fully trained DL models in real-world scenarios, dealing with data that reveals a human's identity poses additional difficulties, as sovereignty over one's data is endangered.
In general, it can be stated that different use cases require different degrees of anonymization to assure human privacy.
On the other hand, different DL models require a different set of undiluted features in order to be able to model the problem at hand.
In the case of facial affective state assessment, most of the context information is unimportant and should be eliminated to reduce the features for re-identification.
Therefore, an approach is needed that offers the research community a pipeline to anonymize faces while only preserving affective state relevant information. 

Further, face anonymization can be vital in promoting ethics and fairness in machine learning.
Not anonymized data can lead to unfair AI decisions, as facial recognition models have been shown to exhibit bias against people of color and women \cite{klare2012face}.
However, current research on face anonymization algorithms often neglects the fact that mere anonymization does not necessarily remove those traits. 
For instance, a face image of a woman of color might still show a woman of color after applying state-of-the-art face anonymization techniques, although her exact identity might not be recognized anymore.
For the task of emotion recognition, in particular, traits like skin color, gender, or hairstyle are not needed, which might introduce bias when being considered.

Additionally, the importance of face anonymization is evident in its ability to protect individual privacy, promote ethical considerations, and ensure compliance with legal requirements. 
By employing face anonymization techniques, researchers can prevent the misuse of personal information and enable the development of machine learning models that are more broadly applicable and ethical.
Face anonymization conceals personal information such as identity, race, ethnicity, gender, or age, reducing the risk of re-identification. 
It is essential in sensitive datasets like medical records and criminal justice data, where anonymity is critical for individuals' privacy and safety. 
It is crucial in healthcare to ensure patient confidentiality when sharing medical images with researchers or medical professionals. 
In the criminal justice system, face anonymization can protect the identity of witnesses, victims, and suspects from potential harm.
The protection of personal data by anonymization or pseudonymization is also enforced in the European Union by law with the General Data Protection Regulation (GDPR) \cite{gruschka2018privacy}. 
Industries such as healthcare and finance are also subject to additional regulations and standards that require anonymization to protect sensitive data. 
For example, US law states that the Health Insurance Portability and Accountability Act (HIPAA) mandates anonymizing Protected Health Information (PHI) to ensure compliance with privacy and security regulations.

To address these shortcomings, this work presents a novel approach to face anonymization that addresses that problem specifically in the context of emotion recognition.
Existing work predominantly tries to find a trade-off between anonymization and task performance by formalizing the problem as a min-max game in which the objective is to find a good compromise between both requirements \cite{nasr2018machine,wu2018towards,wu2019privacy}.
However, features that are neither benefiting the task at hand nor taking away from identity obfuscation (i.e., not affecting either of the two objectives) are mostly ignored.
As such, traits like skin color, gender, or age are still apparent in the anonymized versions of the images, conserving bias and inequity in the underlying data.
Instead of engaging in the aforementioned min-max game, as done by previous approaches, we follow a different paradigm: we completely discard all information except a minimal feature representation that is needed for our chosen use case - emotion recognition - and subsequently re-synthesize arbitrary information for the lost features. 
By doing so, we obtain a complete face image with the same facial expression as the original face while, contrary to existing approaches, removing irrelevant traits for the use case of emotion recognition.
After reviewing relevant literature \cite{ko2018brief,facialemotion2018shivam,emotion2017binh,inpainting2017sun}, we found that facial landmarks can be a good feature set for that task while not exposing too much unnecessary information.
Therefore, as this work focuses on emotion recognition, we chose to extract facial landmarks as an intermediate representation.
To disregard all unimportant information, we chose to extract facial landmarks as an intermediate representation.
Subsequently, we use a Generative Adversarial Network (GAN) architecture, namely \emph{pix2pix} \cite{pix2pix2016}, to re-synthesize a realistic face that incorporates exclusively the features included in the landmarks.
By doing so, our approach - which we call \emph{GANonymization} - has the advantage of not preserving any traits that were not present in the landmark representation. 
As such, features like hairstyle, skin color, or gender are diluted from the intermediate representation, which sets our approach apart from most existing methods.

We evaluate our approach in a three-fold manner:
\begin{enumerate}
    \item We validate if our anonymization method can anonymize faces sufficiently by using a standard measure in this research \cite{serengil2020lightface,serengil2021lightface}.
    \item We validate if our anonymization method keeps important features to preserve emotional expressions by analyzing how the anonymization process affects the predictions of an auxiliary emotion classifier in both a training as well as an inference setting.
    \item We seek to explain the outcomes of the evaluation steps above by analyzing which facial traits are preserved or removed with our anonymization method. To do so, we study how the anonymization process affects the predictions of an auxiliary facial feature detection model.
\end{enumerate}
We show that our approach significantly outperforms state-of-the-art methods in preserving most facial emotional expressions in an anonymized synthesized face.

\section{Related Work}

\begin{figure}[!t]
    \centering
    \includegraphics[width=0.6\columnwidth]{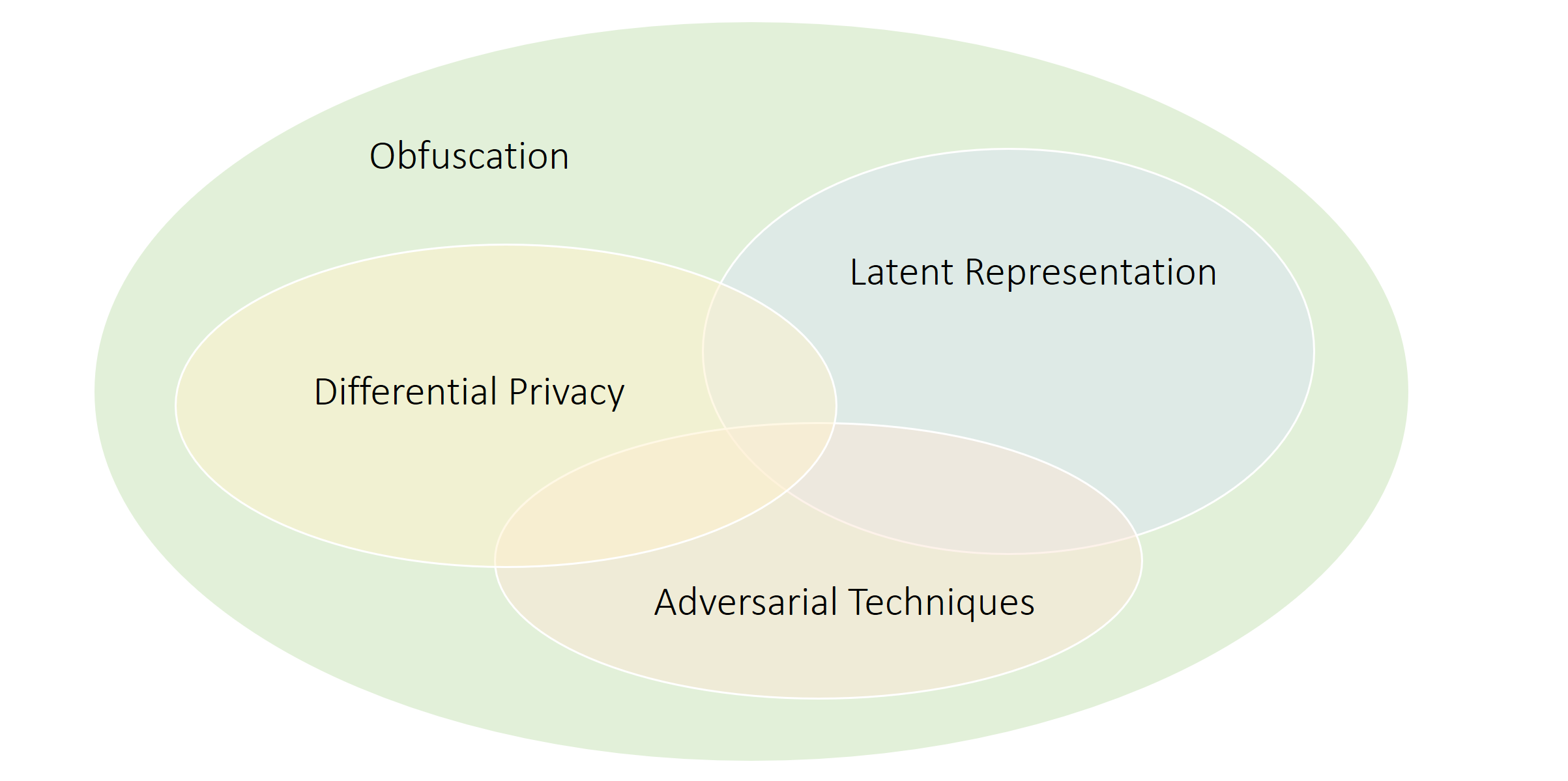}
    \caption{Existing privacy preservation concepts in the context of face anonymization.}
    \label{fig:venn_related_work}
\end{figure}

In this section, we provide an overview of previous research on privacy preservation in the context of facial anonymization. 
The discussion is organized into four key concepts: Obfuscation, Adversarial Techniques, Differential Privacy, and Latent Representations. Note that those concepts are not distinct mechanisms, but different approaches can make use of several of those ideas, as depicted in Figure \ref{fig:venn_related_work}.


\subsection{Obfuscation}
Obfuscation techniques have been pivotal in anonymizing facial data by modifying or masking sensitive areas in images or videos. 
These techniques, including pixelation, blurring, and masking, aim to obscure facial features related to identity while retaining identity-independent characteristics \cite{newton2005preserving}.

For instance, Jourabloo et al. \cite{jourabloo2015attribute} presented an attribute-preserving face de-identification approach. 
While this approach achieved a commendably low face recognition rate, it succeeded in preserving critical facial attributes. 
The method employed an Active Appearance Model and the K-same algorithm to reconstruct new face images while averaging selected features.
Wu et al. \cite{yang2022study} introduced a face-blurring approach to obfuscate faces in the ImageNet dataset, and Raval et al. \cite{raval2017protecting} employed an adversarial perturbation mechanism to protect visual information in camera feeds without substantially impairing application functionality.

Obfuscation techniques are indeed effective in achieving high degrees of anonymity, but they invariably degrade the naturalness and quality of face images, limiting their reusability for diverse facial applications \cite{kuang2021effective}.
In contrast, our approach takes a different path.
Although it involves the removal of various facial traits, it excels in producing high-quality, naturalistic face images. We achieve this by re-synthesizing complete face images using a GAN-based architecture.

\subsection{Adversarial Techniques}
Many existing approaches to facial anonymization are based on training anonymization models using adversarial techniques. 
Generally, the term \emph{adversarial} refers to the paradigm of two \emph{contrary} objectives being maximized at the same time.
For face anonymization, these objectives are the anonymization performance and the so-called \emph{Utility}, i.e., the ability to preserve features that are relevant to solving a certain auxiliary task.
This dual objective can create a min-max game, where improving one objective often results in the degradation of the other.
As such, solving a min-max game with methods of DL inevitably results in finding a compromise between the two objectives.

For example, Nasr et al. \cite{nasr2018machine} developed an adversarial regularization training method aimed at minimizing classification loss while maximizing defense against membership inference attacks. 
Wu et al. \cite{wu2018towards} utilized GANs to learn a degradation transformation that balances action recognition performance with video privacy.
Wu et al. \cite{wu2019privacy} introduced a face de-identification framework that generated de-identified faces with high feature attributes and minimized identity information by incorporating a contrastive verification loss and a structure similarity loss into the GAN training process.

Our approach differs from these methods in that we don't formulate the anonymization problem as a min-max game. 
Instead, we make use of adversarial learning techniques within our framework, particularly by employing a GAN-based architecture to re-synthesize full-face images from our latent representations. 
However, our method stands apart as we don't incorporate \emph{privacy norms} into the GAN training but focus on feature reduction before GAN training. 
This unique approach enables us to remove traits that affect neither anonymization nor utility, setting our method apart from mere compromises between the two.

\subsection{Differential Privacy}
Differential privacy is a concept dependent on the specific application's notion of neighboring databases, which is the core of privacy preservation. 
In deep learning, differential privacy involves the introduction of random noise into a training inference model, which is computed from the underlying stochastic gradient descent (SGD) training gradient. 
This noise is added to ensure a balanced distribution of the results, aligning both utility and privacy considerations \cite{abadi2016deep}.
Complementing differential privacy and the SGD helps maintain a balance between accurate model predictions and privacy protection.

For instance, Croft et al. \cite{croft2021obfuscation} successfully anonymized images by integrating differential privacy into the latent representation of a generative model. 
However, the practical implementation of differential privacy in \emph{real-world scenarios} presents a significant challenge. 
Determining precise privacy boundaries is critical, as adding noise to protect sensitive information may disrupt the entire data distribution, leading to unrecognizable output images \cite{yoon2020anonymization}.

In contrast, our approach does not introduce noise during training or generation. 
Instead, we focus on information reduction before training, retaining only a minimal latent representation, such as facial landmarks. 
While this approach may pose challenges in finding a suitable representation for domains other than emotion recognition, it distinctly sidesteps the pitfalls associated with noisy data distribution and data unrecognizability.

\subsection{Latent Representations}
Traditional GAN-based models often struggle to preserve complex facial attributes, such as emotion, pose, and background, due to image space's high dimensionality and complexity. 
This challenge often results in latent representations being softer in facial style change compared to image space manipulation.
Latent representation, as an abstract and compressed representation inferred from data, captures essential features while discarding redundant information. 
This makes it easier for models to perform tasks like classification and generation. 

Le et al. \cite{le2022styleid} introduced StyleID, a GAN that brings images into a latent representation, uncovers features with significant identity disentanglement, and changes these features in latent space or pixel space. 
However, StyleID may preserve facial traits that have the potential to introduce bias or unfairness, even if they don't correlate directly with identity.
Other methods, such as Sun et al. \cite{inpainting2017sun}, Hu et al. \cite{hu2022protecting}, and Maximov et al. \cite{maximov2020ciagan} with CIAGAN, employed inpainting mechanisms in conjunction with GANs to anonymize faces based on facial landmarks. 
These approaches, while effective, retain context-relevant information outside of the face-segmented area, such as hair color, hairstyle, and gender.
On the other hand, Hukkel\r{a}s and Lindseth introduced DeepPrivacy2 \cite{deepprivacy2}, an enhanced guided GAN framework for anonymizing human figures and faces. 
The DeepPrivacy2 framework entails three detection components for each task: i) face detection with a Dual Shot Face Detector \cite{li2018dsfd}, ii) dense pose estimation with Continuous Surface Embeddings \cite{neverova2020cse}, and iii) Mask R-CNN \cite{he2017maskrcnn} for instance segmentation.
Additionally, three task-specific Surface-guided GANs \cite{hukkelas2022sggan} were trained to synthesize either human figures with conditions, human figures without conditions, or faces.
However, the use of inpainting mechanisms in these approaches may inadvertently retain context-relevant information, potentially introducing bias or unfairness.

In contrast, our approach focuses on excluding context-relevant information by removing all context information except the facial structure with many facial landmarks. 
By concentrating on the elimination of contextual traits, we aim to reduce the potential for bias or unfairness in the dataset.

Overall, DeepPrivacy2 can be regarded as a state-of-the-art full-body anonymization method since it outperformed a variety of other methods in the past \cite{deepprivacy2}.
Furthermore, CIAGAN can be considered as another state-of-the-art face anonymization method, which is also based on landmarks \cite{maximov2020ciagan}.
While CIAGAN utilizes inpainting mechanisms to only anonymize the face area below the forehead, DeepPrivacy2 anonymizes the full facial area, including the forehead.
Consequently, we used DeepPrivacy2 and CIAGAN as the baseline for all our performance evaluations.

\section{Method}

\begin{figure}[!t]
    \centering
    \includegraphics[width=\columnwidth]{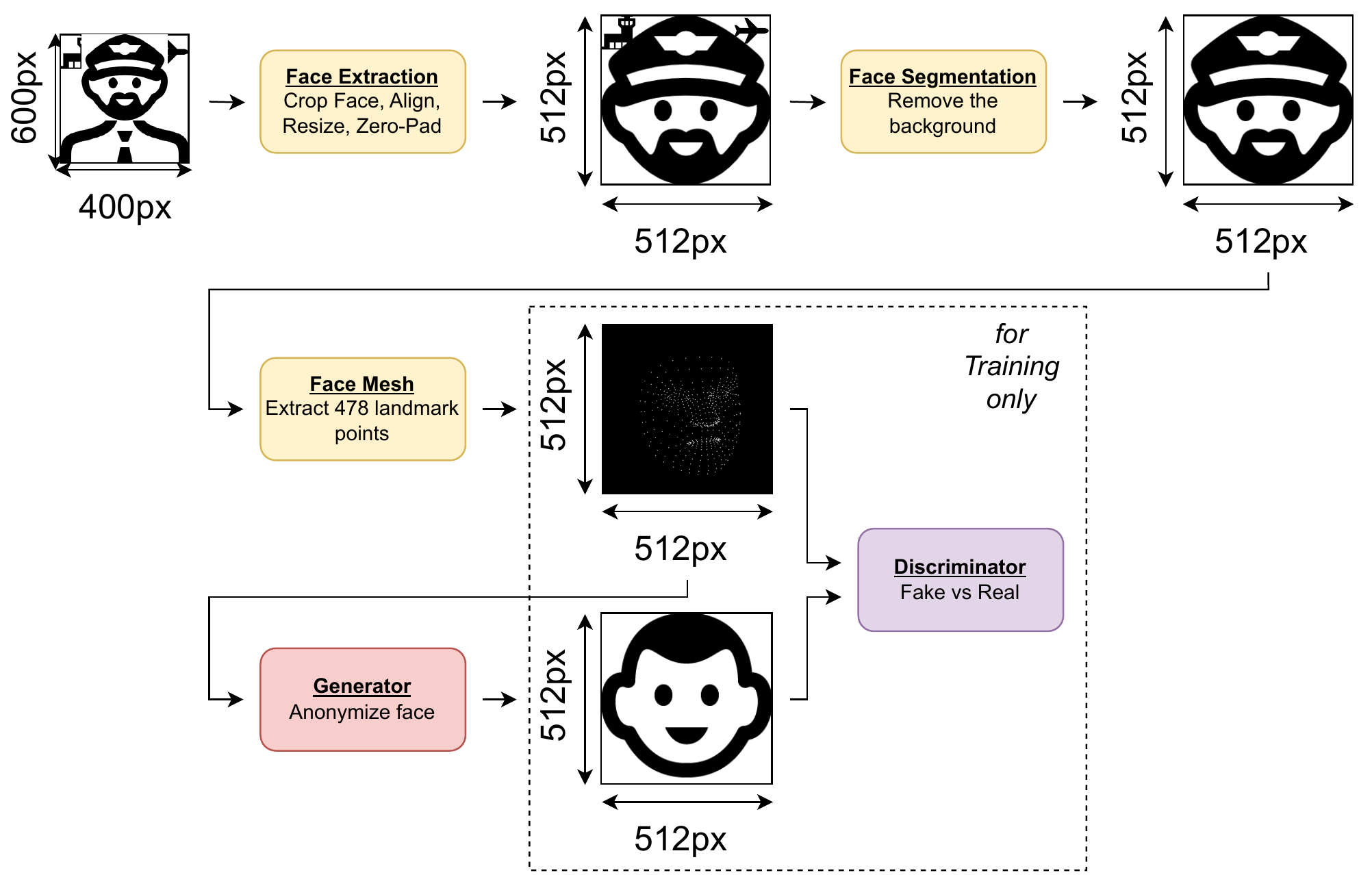}
    \caption{Architecture of the GANonymization pipeline.}
    \label{fig:architecture_ganonymization}
\end{figure}

This section introduces the structure of our GANonymization framework (see Figure~\ref{fig:architecture_ganonymization}) and gives a detailed description of each component and the steps taken for training.\footnote{Our framework's implementation will be made publicly available at \url{https://github.com/hcmlab/GANonymization} upon acceptance.}
The complete GANonymization framework entails four components.

\paragraph{Training Scenario}
In the first step, faces are detected, extracted, and brought into the right format afterward.
The image's background is removed in the second step to eliminate distracting features.
In the third step, facial landmarks are extracted from the face.
In the last step, the GAN's generator synthesizes a new, anonymized face based on those landmarks.
The discriminator evaluates the facial landmarks and the synthesized face to determine whether it is real or fake.

\paragraph{Inference Scenario}
The inference requires fewer steps than the training scenario, as the first and second steps are unnecessary.
Only the extraction of the facial landmarks is required to feed the generator to synthesize an anonymized face.

\subsection{Face Extraction}
\label{subsec:face_extraction}
The first component in the pipeline is face extraction. 
The RetinaFace framework\footnote{https://github.com/serengil/retinaface} \cite{serengil2020lightface} is utilized for this component, which is based on the RetinaFace algorithm \cite{retinaface2020}. 
RetinaFace has been tested against the WIDER \cite{YangWIDER2016} dataset to ensure maximum efficiency in detecting and aligning faces in various scenarios correctly.
However, RetinaFace does not detect all faces every time, especially when factors like poor image quality, extreme angles, or heavy occlusions are in play.
This component includes the following steps:
\begin{enumerate}
    \item \emph{Face Crop.} The input image is analyzed to detect and extract all visible faces.
    \item \emph{Face Align.} According to the literature, aligning the faces supports an increase in accuracy for face recognition models \cite{parkhi2015a}. Therefore, the faces are aligned before the GAN receives them as input. By doing so, the GAN is prevented from focusing too much on the head orientation and instead takes only the face itself into account.
    \item \emph{Image Resize.} The input size of the images for the GAN is set to $512\times512$ pixels. Therefore, the cropped and aligned faces are up-scaled to 512 pixels for the greatest axis, while maintaining the aspect ratio.
    \item \emph{Zero Padding.} To achieve the final $512\times512$ pixels for the required input shape of the GAN, we apply zero padding to the sides [(right and left) or (top and bottom)] of the image to keep the face centered in the image.
\end{enumerate}

\subsection{Face Segmentation}
\label{subsec:face_segmentation}
The second component of the pipeline is face segmentation. 
Even though this step could be skipped, we observed that the pix2pix architecture we used for re-synthesis of the faces (see Section \ref{subsec:resynthesis}) yielded visually better results when not having to deal with variations in the background. 
Consequently, the original background is removed by applying face segmentation and setting all pixel intensities outside the face segments to $0$. 
Therefore, a head segmentation model\footnote{https://github.com/wiktorlazarski/head-segmentation} based on a U-Net is utilized.

\subsection{Facial Landmarks}
After the pre-processing steps, we generate intermediate representations of the faces. 
Here, we aim for a representation that (i) does not contain information that could be used to identify the original face and (ii) holds all necessary information needed for facial expression analysis tasks.
Existing literature on the topic \cite{ko2018brief,facialemotion2018shivam,emotion2017binh,inpainting2017sun} indicates that facial landmarks fulfill both of these requirements in the context of emotion recognition.
Note that although this work focuses on the context of emotion recognition exclusively, the concept could be transferred to other domains as well.
Therefore, a suitable intermediate representation, which might not be facial landmarks, would have to be found for the specific task.
For our experiments, we extract 478 3-dimensional coordinate facial landmarks utilizing the media-pipe face-mesh model \cite{kartynnik2019real} to receive an abstract representation of the facial shape.
The resulting 3D landmarks are projected onto a 2D image with a black background where each landmark point is represented by a single white pixel.
It is necessary to translate the 3D landmarks into a 2D image due to the image-to-image type of model used for the re-synthesis of the faces (as described in the following section~\ref{subsec:resynthesis}).

\subsection{Re-Synthesis}
\label{subsec:resynthesis}
To obtain an anonymized version of the input that still looks highly realistic, we aim for a re-synthesis of high-quality faces.
Therefore, we use the \emph{pix2pix} architecture, a GAN-based image-to-image model.
The original purpose of \emph{pix2pix} is to convert image data from a particular source domain to a target domain.
Our specific goal in the re-synthesis stage is to transfer the landmark representations back to random, high-quality face images that expose the same facial landmark structure.
The \emph{pix2pix} architecture has been successfully applied to similar use cases in the past, e.g., for synthetic data augmentation in the context of defect detection \cite{DBLP:pix2pixCARBON1,mertes2020alternative}, where segmentation masks of material defects (which, on a technical level, are quite similar to visual landmark representations) were converted to realistic looking data.
More recent GAN-based architectures like ProGAN \cite{DBLP:progan}, StyleGAN \cite{DBLP:stylegan}, or StyleGANv2 \cite{DBLP:stylegan2}, that impress with their ability to generate hyper-realistic data, are specifically designed to create new data from scratch. To use those models for image-to-image conversion tasks, a projection of the input image has to be found in the GAN's latent space, which is highly inefficient and might not be possible at all for some data instances. As such, we chose to use \emph{pix2pix}, as it is specifically tailored for end-to-end image-to-image translation. 
For the training of the \emph{pix2pix} model, we used existing face images as the \emph{target} domain, whereas for the \emph{source} domain, we used landmark representations that we priory extracted from those images. 
In other terms, we trained the \emph{pix2pix} network to learn the inverse transformation of a landmark extractor - we perform an image-to-image translation from an image representation of landmark features to realistic-looking face images. 
By using that approach, we are able to automatically create geometrically aligned source/target image pairs for training.
Contrary to architectures such as CycleGAN \cite{DBLP:conf/iccv/ZhuPIE17} that work with non-parallel training data, \emph{pix2pix} directly takes advantage of having mapped training pairs, which again supports our architecture choice.

We process the CelebA \cite{liu2015faceattributes} dataset within our pipeline to extract and align the faces (section \ref{subsec:face_extraction}), remove the background of the faces by face segmentation (section \ref{subsec:face_segmentation}), and extract a face-mesh of each face which represents the landmark/image pairs for training.
CelebA was used because of its size (202,599 images) and because it contains only images of high quality - using low-quality images would limit the quality of GANonymization's output images unnecessarily.
We used the same pipeline for the landmark extraction to anonymize the images. 
Additionally, training images were normalized to $mean=(0.5, 0.5, 0.5)$ and $std=(0.5, 0.5, 0.5)$.
Our implementation was built upon Erik Linder-Norén's pix2pix implementation\footnote{\url{https://github.com/eriklindernoren/PyTorch-GAN\#pix2pix}}, which in turn strongly adheres to the original \emph{pix2pix} publication\cite{pix2pix2016}.
We trained the model for 25 epochs with a batch size of 32. 
The Adam optimizer was used with a learning rate of 0.0002, $\beta_1$ decay of 0.5, and $\beta_2$ decay of 0.999.
After training, our model could transfer landmark representations to face images that show the same facial expression expressed by the original face.
In the case of an issue with face detection and, therefore, no available facial landmarks, an empty (black) image can be inferred with our model with the result of a synthesized average face, which is based on the faces seen by the model during the training process.
Exemplary outputs of our pipeline are shown in Figures~\ref{fig:wider_synthesized_samples},~\ref{fig:affectnet_synthesized_samples},~\ref{fig:ckplus_synthesized_samples},~\ref{fig:faces_synthesized_samples},~and~\ref{fig:celeba_synthesized_samples}.

\section{Evaluation}
In the following sections, we describe how we validate our approach using three different evaluations. 
First, we evaluate the anonymization capability of the approach. 
Second, we evaluate the suitability of the approach for the task of emotion recognition, i.e., whether our approach preserves information that is relevant to facial emotion recognition. 
Finally, we go into detail about the facial features that get preserved or removed with our anonymization approach.

\subsection{Anonymization Performance}
\label{subsec:anonymization_performance}
In this first part of the evaluation, the anonymization performance of our approach was assessed.
Hereby, with the term anonymization performance, we refer to the capability of the method to alter input images in a way that they ideally cannot be re-identified.
Therefore, we compared the synthesized images of our approach with the original images and versions synthesized by DeepPrivacy2, CIAGAN, and basic methods like pixelation and blurring.

\subsubsection{Dataset}
The dataset used for the comparison was the WIDER \cite{YangWIDER2016} dataset, which is commonly used for benchmarking face detection algorithms. 
Further, the authors of DeepPrivacy2 had already used it in their original publication. 
Therefore, by using it in our experiments too, we do not introduce a bias towards GANonymization by using a dataset that DeepPrivacy2 might not be suited for.
It contains images of people in various sceneries whose faces vary in scale, pose, and occlusion. 
In each image, one or more faces are apparent.
In total, WIDER embodies 32,203 images in 61 event settings. 
The many different head orientations, obfuscations, facial expressions, lighting conditions, and others enable an optimal evaluation setting to measure the overall performance in anonymizing these faces. 
After we applied our pre-processing pipeline with the face extraction (section~\ref{subsec:face_extraction}) and face segmentation (section~\ref{subsec:face_segmentation}) components, the images were split into a training and validation set of 92,749 and 22,738 face images, respectively.

\subsubsection{Setup}
\begin{figure}[!ht]
    \centering
    \includegraphics[width=1\textwidth]{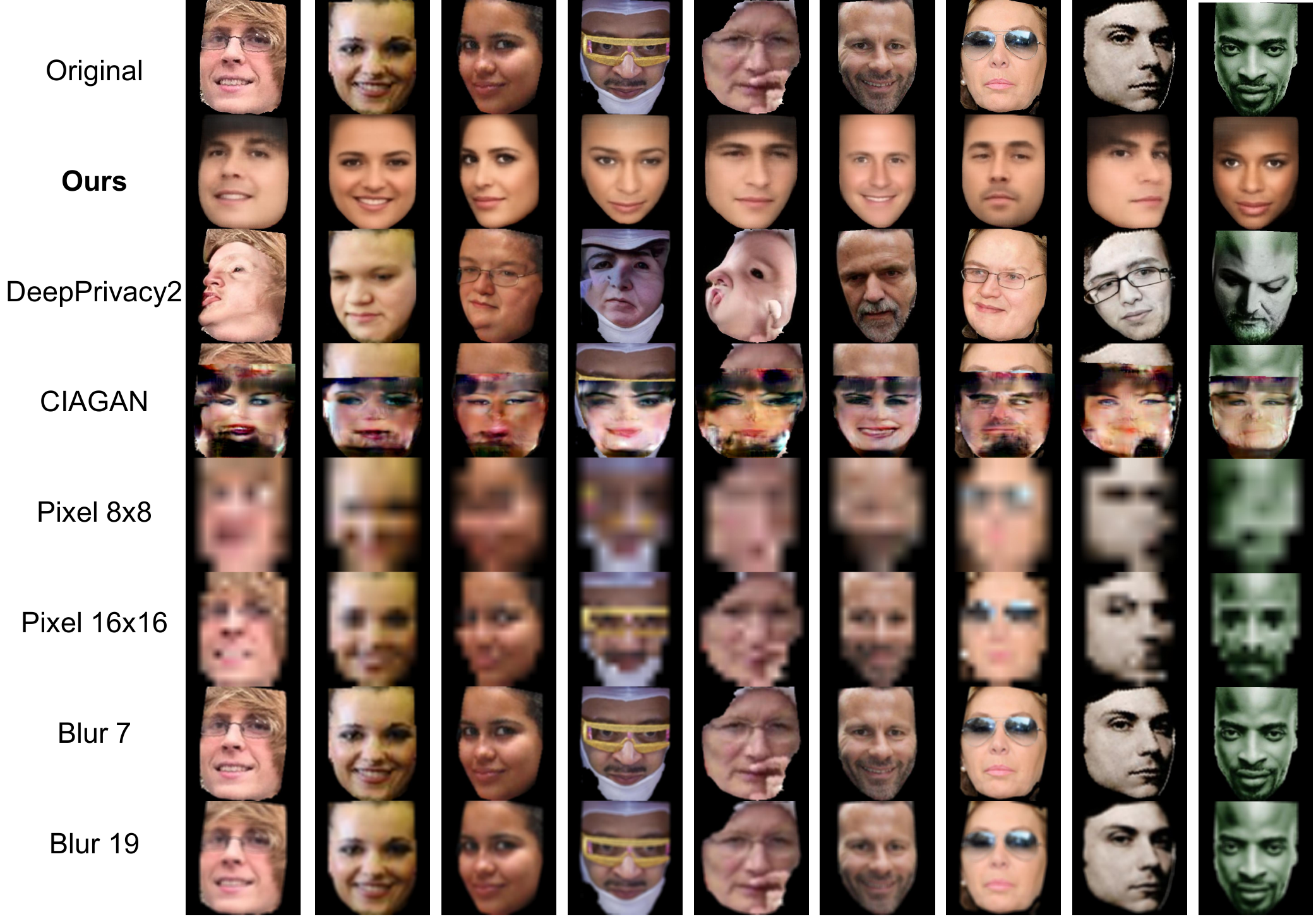}
    \caption{Sample of synthesized faces based on the WIDER dataset.}
    \label{fig:wider_synthesized_samples}
\end{figure}

The performance measurement is based on the comparison of the original images and their synthesized counterparts. The synthesized images are produced by our method, DeepPrivacy2, and CIAGAN, respectively.
Exemplary anonymized images for WIDER can be seen in Figure~\ref{fig:wider_synthesized_samples}.

\subsubsection{Metric}
\label{subsubsec:metric}
A widely used method to assess the anonymization degree of a face image is to compute the cosine distance between image encodings of the original and anonymized image versions. 
Here, a lower cosine distance equals higher similarity between the faces and is commonly considered as the anonymized face being \emph{more recognizable} to the original face.
Specialized frameworks for face recognition like DeepFace\footnote{https://github.com/serengil/deepface} make use of that paradigm and thus can be used as an evaluation tool for anonymization algorithms \cite{serengil2020lightface,serengil2021lightface}.
As such, for the comparison of the anonymization performance of our approach versus the other methods, we use the DeepFace framework.
As a backbone model for image encoding, we use the state-of-the-art face recognition model Facenet512 \cite{firmansyah2023comparison}, which is also integrated into DeepFace.
The cosine distance is defined as follows:
\begin{equation}
    cdistance = 1 - \frac{I_o \cdot I_a}{\lVert I_o \rVert \lVert I_a \rVert}
\end{equation}
where $I_o$ and $I_a$ are the Facenet512 feature embedding space representations of the original and anonymized images, respectively.
When the cosine distance exceeds $0.3$, it indicates that the feature embedding space has diverged significantly from the original space, making re-identification impractical.
We computed the cosine distance of the image pairs for each method with the original image.

\subsubsection{Results}

\begin{table}
    \centering
    \begin{tabular}{|l|c|}
        \toprule
        \textbf{Method} & \textbf{Cosine Distance} \\
        \midrule
        Original & 0.0000 \\
        \textbf{Ours} & \textbf{0.7145} \\
        DeepPrivacy2\cite{deepprivacy2} & \textbf{0.8119} \\
        CIAGAN\cite{maximov2020ciagan} & \textbf{0.9280} \\
        Pixel 8x8 & \textbf{0.8791} \\
        Pixel 16x16 & \textbf{0.6651} \\
        Blur 9x9 & 0.0102 \\
        Blur 17x17 & 0.0725 \\
        \bottomrule
    \end{tabular}
    \caption{The mean cosine distances between the original images and the anonymized versions obtained through GANonymization, DeepPrivacy2 (DP2), CIAGAN (CIA), pixelation with a kernel sized 8x8 and 16x16, and blurring with a kernel sized 9x9 and 17x17. The methods with a cosine distance in bold exceed the threshold of $0.3$.}
    \label{tab:anonym_perform_cosine}
\end{table}

Our approach achieved a mean cosine distance of $0.7145$, while DeepPrivacy2 and CIAGAN reached a greater cosine distance of $0.8119$ and $0.9280$, respectively (see Table \ref{tab:anonym_perform_cosine}).
The pixelation with a kernel sized $8\times8$ achieved $0.8791$, while the bigger kernel sized $16\times16$ achieved $0.6651$.
Blurring with a kernel sized $9\times9$ and $17\times17$ stayed below the threshold necessary for no re-identification with a cosine distance of $0.0102$ and $0.0725$, respectively.

\subsubsection{Discussion}
Our evaluation measures the mean cosine distance between the Facenet512-face-based image encodings of original and anonymized face images.
Accordingly, the distance between two encodings marks the non-similar features and how complex the reconstruction of one encoding towards another encoding is, which is conventionally interpreted as \emph{degree of anonymization}.
Comparing the results of our approach with the others, we found that DeepPrivacy2, CIAGAN, and pixelation achieved a mean cosine distance above the threshold of $0.3$, which indicates that the feature embedding space diverged significantly from the original image.

While pixelation changes only the underlying image resolution to obfuscate the face, the quality of the image suffers accordingly and the face could still be re-identified - at least for the kernel sized $16\times16$.
Blurring does not modify the image resolution but reduces the overall image quality nonetheless.

On the other hand, CIAGAN synthesized a new face inside of the facial landmark segment of the original image.
The result of the synthesized face inside the original image by CIAGAN lacks in quality.
However, a face with its emotional expression can still be determined.
The low quality and high number of artifacts can be a reason for the high cosine distance to the original image.

DeepPrivacy2, on the other side, synthesized a face that does not necessarily preserve the orientation of the face or the facial expression.
In some cases, it can be observed that the outputted face does not have much similarity to a face due to the extreme dysmorphism of facial areas.
Accordingly, the dysmorphism can be a result of the increased cosine distance to the original images compared to our approach.

Therefore, we can claim that our approach has a great quality in synthesized faces and solid anonymization performance despite the lower cosine distance compared to DeepPrivacy2, CIAGAN, and pixelation $8\times8$.

\subsection{Preserved Emotional Expressions}
\label{subsec:preserved_facial_expressions}
After showing our approach's anonymization capabilities in section \ref{subsec:anonymization_performance}, we need to ensure that this performance does not come at the expense of the primary task that the data will be used for, in our case, affect recognition. Thus, in this section, we examine whether our method can anonymize faces while maintaining their original emotional expressions.
For this evaluation, we use three different datasets which are commonly used in the research field of affect recognition, namely \emph{AffectNet} \cite{affectnet2017mollahosseini}, \emph{CK+} \cite{CkPlus2010Patrick}, and \emph{FACES} \cite{ebner2010faces}.

\subsubsection{Datasets}
We used three different datasets to cover a wide variety of different settings. 

The first dataset we've chosen is the \emph{AffectNet} dataset.
We chose it because it contains in-the-wild data, resulting in emotions being expressed in a quite natural way.
It contains around 0.4 million images manually labeled according to eight emotional expressions: \emph{Neutral}, \emph{Happy}, \emph{Angry}, \emph{Sad}, \emph{Fear}, \emph{Surprise}, \emph{Disgust}, and \emph{Contempt}. 
The faces in this dataset have a great variety of individuals, head orientations, lighting conditions, and ethnicities. 
The dataset was pre-processed with face extraction (section \ref{subsec:face_extraction}) and face segmentation (section~\ref{subsec:face_segmentation}). 
In the process, images in which no face was detected were discarded.
Accordingly, the training and validation splits contained 210,174 and 2,874 images, respectively.

The second dataset, namely \emph{CK+}, contains 593 video sequences with 123 subjects aged 18 to 50 years and of various genders and heritage. 
Each video sequence shows the transition from a neutral facial expression to a non-neutral one, recorded at 30 frames per second.
We chose the dataset because, due to the emotional transitions, single image frames also cover facial expressions where the emotions are shown quite subtly.
Overall, 327 of those videos are labeled with one of seven emotional expressions: \emph{Anger}, \emph{Contempt}, \emph{Disgust}, \emph{Fear}, \emph{Happiness}, \emph{Sadness}, and \emph{Surprise}. 
Again, we applied our pre-processing pipeline with face extraction and face segmentation on the dataset and received a training and validation set of 259 and 68 images, respectively.

Lastly, the \emph{FACES} dataset with a total of 2,052 images with different age groups and gender embodies six emotional expressions: \emph{Neutral}, \emph{Sad}, \emph{Disgust}, \emph{Fear}, \emph{Anger}, and \emph{Happy}. 
We used that dataset as it contains only images of acted emotions, making it a good counterpart for the other two datasets. 
By including it, we also cover emotional expressions that are shown in a rather exaggerated way.
The images in this dataset have high quality.
Further, the dataset contains only frontal shots of the faces with optimal lighting conditions. 
As was done for the previous datasets, we also applied pre-processing with face extraction and face segmentation, resulting in 1,827 images in the training split and 214 images in the validation split.

\subsubsection{Setup}
\label{subsubsec:emotion_setup}
\begin{figure}[!ht]
    \centering
    \includegraphics[width=1\textwidth]{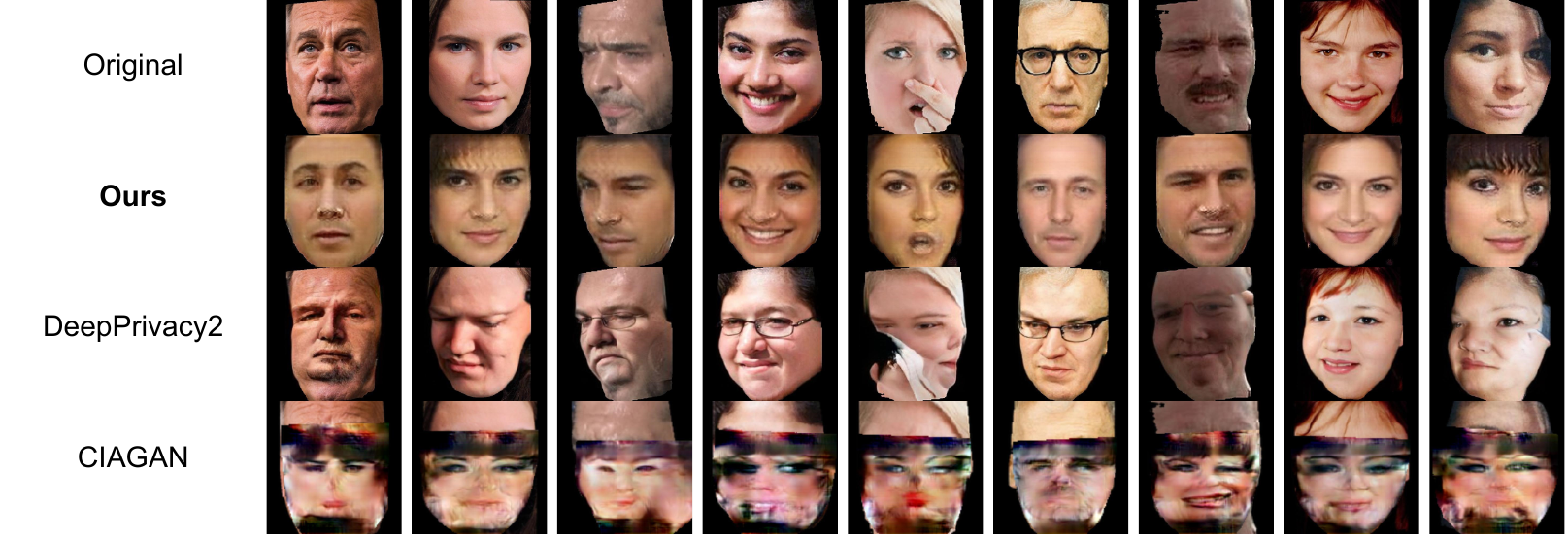}
    \caption{Sample of synthesized faces based on the AffectNet dataset.}
    \label{fig:affectnet_synthesized_samples}
\end{figure}

\begin{figure}[!ht]
    \centering
    \includegraphics[width=1\textwidth]{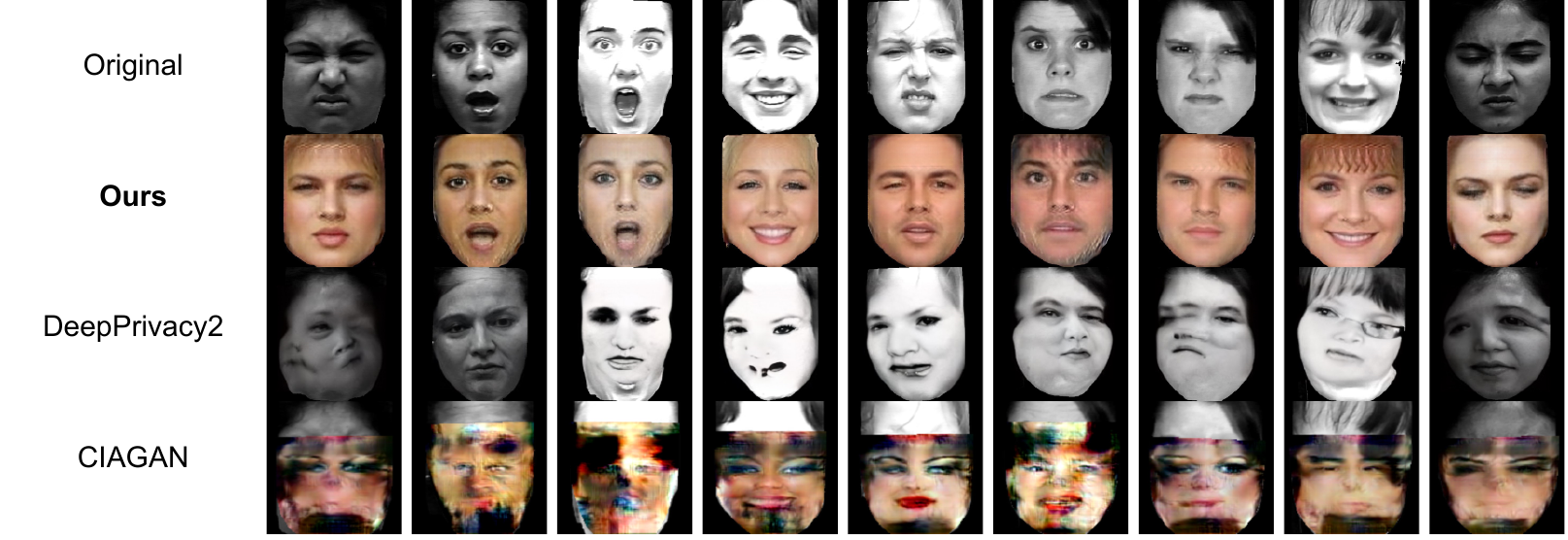}
    \caption{Sample of synthesized faces based on the CK+ dataset.}
    \label{fig:ckplus_synthesized_samples}
\end{figure}

\begin{figure}[!ht]
    \centering
    \includegraphics[width=1\textwidth]{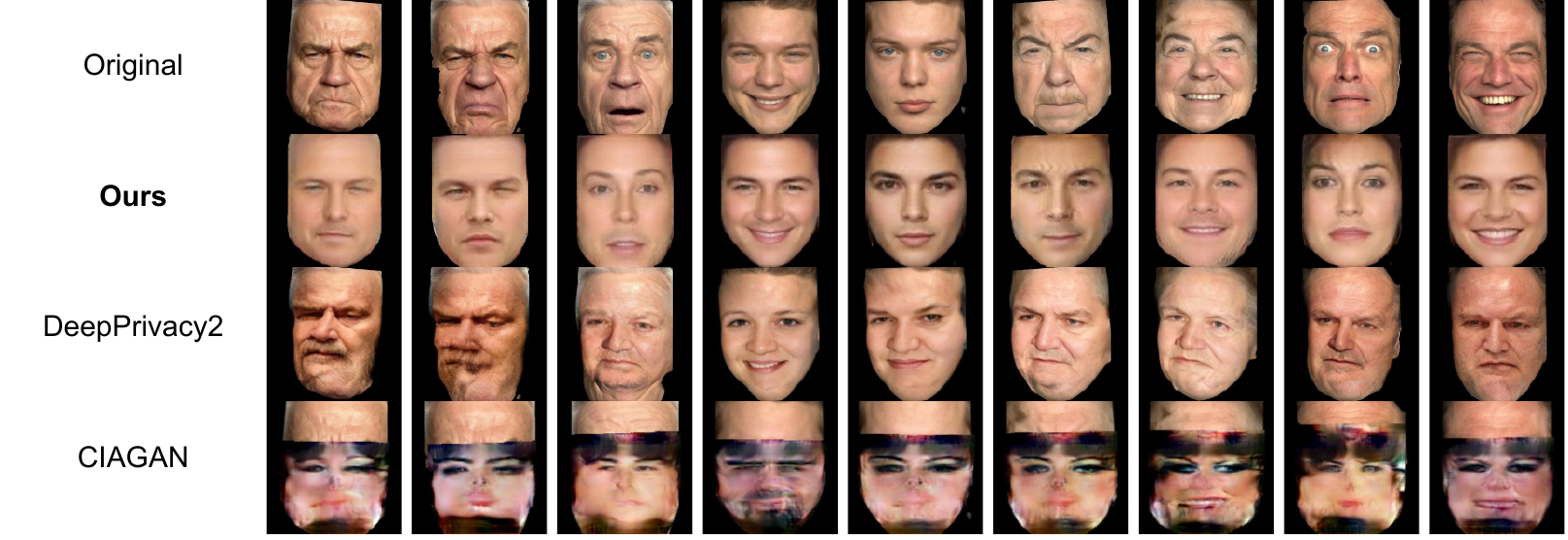}
    \caption{Sample of synthesized faces based on the FACES dataset.}
    \label{fig:faces_synthesized_samples}
\end{figure}

We created anonymized versions of the three datasets, resulting in 12 datasets in total: the three original ones, those anonymized with GANonymization, and those anonymized with DeepPrivacy2 and CIAGAN.
Exemplary anonymized images for AffectNet, CK+, and FACES can be seen in Figure~\ref{fig:affectnet_synthesized_samples},~\ref{fig:ckplus_synthesized_samples},~and~\ref{fig:faces_synthesized_samples}, respectively.
Note that although the CK+ dataset consists of greyscale images, the anonymized versions of our approach are colored - this is a nice byproduct of our approach since we only use the landmarks as an intermediate representation, whereas the re-synthesis is still based on the GAN that was trained on CelebA.
We splitted the evaluation of the emotional expression preserving capabilities into two sub-evaluations.
First, we assessed how the emotional expression gets preserved during an \emph{inference} setting, thus, how a model trained on original data behaves when fed with anonymized data.
Second, we evaluated how the model influences the training process of a model trained on anonymized data.

\paragraph{Inference Scenario Evaluation.}

To measure how well GANonymization can preserve emotional expressions, we first trained an emotion classifier separately for the three original datasets.
Subsequently, we applied the trained models to the original and the anonymized datasets and studied the prediction changes caused by the anonymization methods.
Here, big changes in prediction probability can be interpreted as poor preservation of features contributing to emotional expressions.
We decided to go for three separate dataset-specific models instead of one across-dataset model, as our evaluation methodology relies on the classifiers accurately modeling the relation between data and emotion for the specific datasets.
As the datasets differ substantially, we argue that an across-dataset model, although having the potential to gain a greater overall generalizability, would under-perform on the single datasets due to dataset-specific details that would get lost (e.g., the CK+ dataset is greyscale, FACES are frontal-only, etc.).

As classifier architecture, we chose the base version of the ConvNeXt, which is considered one of the state-of-the-art DL architectures for computer vision tasks \cite{LiuConvNeXt2022}. 
Furthermore, the model was pre-trained on the ImageNet \cite{DengImageNet2009} dataset. 
The classification model's last linear layer's amount of output nodes was changed to match the number of classes, which differed for each dataset.
We used the cross-entropy loss for training.
Class weights were calculated on the train split of each dataset individually.
The AdamW \cite{IlyaAdamW2017} optimizer was used with a learning rate of $0.0003$ and a weight decay of $0.001$.
Additionally, the learning rate was reduced when the validation loss reached a plateau for three consecutive epochs.
The images were pre-processed by normalizing with $mean=(0.485, 0.456, 0.406)$ and $std=(0.229, 0.224, 0.225)$ for both, training and testing.
Hereby, the mean and standard values for normalization were based on the pre-trained model's dataset (ImageNet).
During the training phase, images were randomly flipped horizontally with a probability of 50\% for data augmentation.
The classification models converged on the validation split within 3, 12, and 9 epochs for AffectNet, CK+, and FACES, respectively. 
For comparing the anonymization approaches, namely ours, DeepPrivacy2, and CIAGAN, we used the trained emotion classifiers to make predictions on the original images as well as for the anonymized versions.
By doing so, we can assess to which degree the anonymization process preserves features that hold information on emotional expressions.

\paragraph{Training Scenario Evaluation.}
In this sub-evaluation, we assess how the performance of an emotion recognition model's performance degraded when trained on the anonymized versions.
To do so, we used the same classifiers that were trained in the \emph{Inference Scenario} but additionally trained the same architecture once with the data anonymized by GANonymization and once anonymized by DeepPrivacy2 and CIAGAN. 
Thus, we use 12 different models for this experiment, each trained on one of the datasets mentioned above.
Subsequently, we compare the performance of the models on the original datasets' validation splits.

\subsubsection{Metric}

\paragraph{Inference Scenario Evaluation.}
We measure the ability of each anonymization approach to preserve the original emotional expressions by looking at how the prediction probabilities for the emotion classifiers change when applied to the original datasets vs. each of the anonymized datasets.
I.e., for each image, we measure how the class probability of a certain emotion predicted from the original image differs from the class probability of that same emotion in the anonymized version of the image. 
Subsequently, we average the resulting probability differences of the images in the validation sets for each emotion.
Here, a higher mean difference indicates that the anonymization process obfuscated more features defining the respective emotion.
In comparison, a lower difference implies that the anonymization process preserved more emotion-related features.

\paragraph{Training Scenario Evaluation.}
Here, we compare the F1 score of the different models on the respective validation splits.
F1 score was chosen as, especially in the CK+ data, a relatively high class imbalance is apparent.

\subsubsection{Results}
\begin{figure}[!ht]
    \centering
    \subfigure[AffectNet]{\includegraphics[width=0.3\columnwidth]{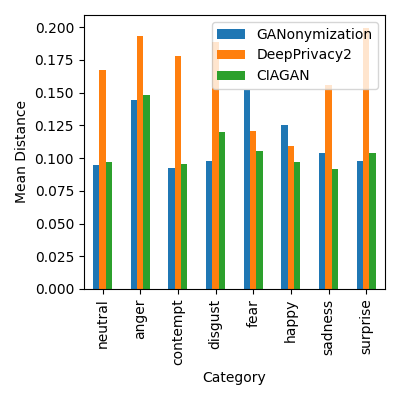}}
    \subfigure[CK+]{\includegraphics[width=0.3\columnwidth]{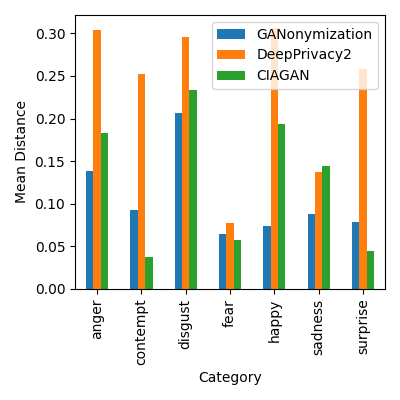}}
    \subfigure[FACES]{\includegraphics[width=0.3\columnwidth]{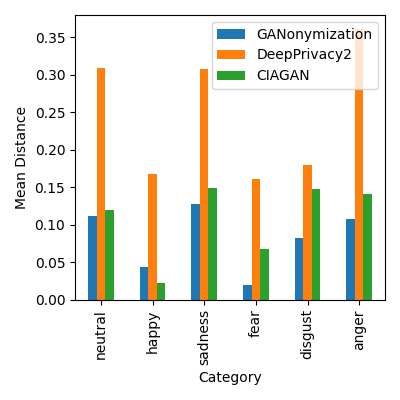}}
    \caption{The mean distance of the class probability prediction for each emotion for our method, DeepPrivacy2, and CIAGAN on each dataset (lower is better).}
    \label{fig:emotion_classification_mean}
\end{figure}

\paragraph{Inference Scenario Evaluation.}
The results are depicted in Figure \ref{fig:emotion_classification_mean} and Table \ref{tab:emotion_classification_mean}.
As can be seen, GANonymization outperformed DeepPrivacy2 in all emotions except \emph{Fear} and \emph{Happy} in the AffectNet dataset.
Compared to CIAGAN, our approach outperformed in most emotions except \emph{Fear}, \emph{Happy}, and \emph{Sadness} in the AffectNet dataset, also \emph{Contempt}, \emph{Fear}, and \emph{Surprise} in the CK+ dataset, and only \emph{Happy} in the FACES dataset.

\begin{table}[!ht]
    \centering
    \subfigure[AffectNet]{
        \renewcommand{\arraystretch}{0.9} 
        \setlength{\tabcolsep}{3pt}
        \begin{tabular}{lrrr}
        \toprule
        {} &  Ours &  DP2 &    CIA \\
        \midrule
        Neutral  &        \textbf{0.09} &      0.17 &  0.10 \\
        Anger    &        \textbf{0.14} &      0.19 &  0.15 \\
        Contempt &        \textbf{0.09} &      0.18 &  0.10 \\
        Disgust  &        \textbf{0.10} &      0.19 &  0.12 \\
        Fear     &        0.15 &      0.12 &  \textbf{0.11} \\
        Happy    &        0.13 &      0.11 &  \textbf{0.10} \\
        Sadness  &        0.10 &      0.16 &  \textbf{0.09} \\
        Surprise &        \textbf{0.10} &      0.20 &  0.10 \\
        \bottomrule
        \end{tabular}
    }
    \subfigure[CK+]{
        \renewcommand{\arraystretch}{0.9} 
        \setlength{\tabcolsep}{3pt} 
        \begin{tabular}{lrrr}
        \toprule
        {} &  Ours &  DP2 &    CIA \\
        \midrule
        & & & \\
        Anger    &        \textbf{0.14} &      0.30 &  0.18 \\
        Contempt &        0.09 &      0.25 &  \textbf{0.04} \\
        Disgust  &        \textbf{0.21} &      0.30 &  0.23 \\
        Fear     &        0.06 &      0.08 &  \textbf{0.06} \\
        Happy    &        \textbf{0.07} &      0.31 &  0.19 \\
        Sadness  &        \textbf{0.09} &      0.14 &  0.14 \\
        Surprise &        0.08 &      0.26 &  \textbf{0.05} \\
        \bottomrule
        \end{tabular}
    }
    \subfigure[FACES]{
        \renewcommand{\arraystretch}{0.9} 
        \setlength{\tabcolsep}{3pt}
        \begin{tabular}{lrrr}
        \toprule
        {} &  Ours &  DP2 &    CIA \\
        \midrule
        Neutral &        \textbf{0.11} &      0.31 &  0.12 \\
        Anger   &        \textbf{0.11} &      0.36 &  0.14 \\
        & & & \\
        Disgust &        \textbf{0.08} &      0.18 &  0.15 \\
        Fear    &        \textbf{0.02} &      0.16 &  0.07 \\
        Happy   &        0.04 &      0.17 &  \textbf{0.02} \\
        Sadness &        \textbf{0.13} &      0.31 &  0.15 \\
        & & & \\
        \bottomrule
        \end{tabular}
    }
    \caption{The mean class probability distances between the original images and the anonymized versions obtained through GANonymization, DeepPrivacy2 (DP2), and CIAGAN (CIA).}
    \label{tab:emotion_classification_mean}
\end{table}

To assess if these differences are statistically significant, we conducted statistical hypothesis tests for each emotion as well as each dataset.
As a Shapiro-Wilk test revealed that the data was not normally distributed for any of the datasets, Wilcoxon tests were used for the post-hoc analysis.
Subsequently, we did a dataset-wise p-value correction using Bonferroni's method. 
We report the resulting statistics in Table \ref{tab:emotion_pvalues}.
As can be seen, we found significant differences for all emotions in the AffectNet dataset for DeepPrivacy2 and CIAGAN except for \emph{Neutral} and \emph{Anger}.
In CK+, we found significant differences for all classes except \emph{Sadness} and \emph{Surprise} for DeepPrivacy2 and \emph{Disgust} for CIAGAN.
In the FACES dataset, we found significant differences for all classes except \emph{Happy} and \emph{Fear} for DeepPrivacy2 and \emph{Happy} and \emph{Anger} for CIAGAN.

\begin{table}[]
    \centering
    \subfigure[AffectNet]{
        \begin{tabular}{l|rrr|rrr|r}
        \toprule
        {} & \multicolumn{3}{|c|}{Ours vs. DeepPrivacy2} & \multicolumn{3}{|c|}{Ours vs. CIAGAN} & {} \\
        {} &        $p$ &  $Z$ &         $r$ &        $p$ &  $Z$ &         $r$ &     $N$ \\
        \midrule
        neutral  &  <0.001\textbf{***} &   -26.149391 & -0.499194 &  0.119 &   -26.183915 & -0.492635 &  2744 \\
        anger    &   <0.001\textbf{***} &   -10.844502 & -0.207023 &   1.000 &   -10.331517 & -0.194381 &  2744 \\
        contempt &  <0.001\textbf{***} &   -26.046682 & -0.497233 &  0.013\textbf{*} &   -26.187686 & -0.492706 &  2744 \\
        disgust  &  <0.001\textbf{***} &   -25.431412 & -0.485488 &  <0.001\textbf{***} &   -25.441451 & -0.478666 &  2744 \\
        fear     &   <0.001\textbf{***} &   -15.905089 & -0.303630 &   <0.001\textbf{***} &   -15.850979 & -0.298227 &  2744 \\
        happy    &   <0.001\textbf{***} &   -12.989478 & -0.247970 &   <0.001\textbf{***} &   -12.821720 & -0.241233 &  2744 \\
        sadness  &   <0.001\textbf{***} &    -4.724173 & -0.090185 &   <0.001\textbf{***} &    -4.195525 & -0.078936 &  2744 \\
        surprise &  <0.001\textbf{***} &   -25.921903 & -0.494851 &   0.034\textbf{*} &   -26.142871 & -0.491863 &  2744 \\
        \bottomrule
        \end{tabular}
    }
    \subfigure[CK+]{
        \begin{tabular}{l|rrr|rrr|r}
        \toprule
        {} & \multicolumn{3}{|c|}{Ours vs. DeepPrivacy2} & \multicolumn{3}{|c|}{Ours vs. CIAGAN} & {} \\
        {} &        $p$ &  $Z$ &         $r$ &        $p$ &  $Z$ &         $r$ &     $N$ \\
        \midrule
        anger    &  <0.001\textbf{***} &    -3.941178 & -0.477938 &  <0.001\textbf{***} &    -3.415688 & -0.414213 &  68 \\
        contempt &  <0.001\textbf{***} &    -4.326130 & -0.524620 &  <0.001\textbf{***} &    -6.495306 & -0.787672 &  68 \\
        disgust  &  <0.001\textbf{***} &    -3.635660 & -0.440889 &  0.158 &    -1.411492 & -0.171169 &  68 \\
        fear     &  0.002\textbf{**} &    -3.067397 & -0.371977 &  0.001\textbf{**} &    -3.201825 & -0.388278 &  68 \\
        happy    &  <0.001\textbf{***} &    -3.415688 & -0.414213 &  <0.001\textbf{***} &    -3.440129 & -0.417177 &  68 \\
        sadness  &  0.146 &    -1.454264 & -0.176355 &  <0.001\textbf{***} &    -4.741634 & -0.575008 &  68 \\
        surprise &  0.051 &    -1.949203 & -0.236376 &  <0.001\textbf{***} &    -4.069495 & -0.493499 &  68 \\
        \bottomrule
        \end{tabular}
    }
    \subfigure[FACES]{
        \begin{tabular}{l|rrr|rrr|r}
        \toprule
        {} & \multicolumn{3}{|c|}{Ours vs. DeepPrivacy2} & \multicolumn{3}{|c|}{Ours vs. CIAGAN} & {} \\
        {} &        $p$ &  $Z$ &         $r$ &        $p$ &  $Z$ &         $r$ &     $N$ \\
        \midrule
        neutral &  <0.001\textbf{***} &    -6.869167 & -0.469567 &  <0.001\textbf{***} &    -4.080480 & -0.278936 &  214 \\
        happy   &  0.965 &    -0.043556 & -0.002977 &  0.102 &    -1.633626 & -0.111672 &  214 \\
        sadness &  <0.001\textbf{***} &    -4.305428 & -0.294313 &  0.018\textbf{*} &    -2.366910 & -0.161799 &  214 \\
        fear    &  0.287 &    -1.064641 & -0.072777 &  <0.001\textbf{***} &    -8.291628 & -0.566804 &  214 \\
        disgust &  0.014\textbf{*} &    -2.459535 & -0.168130 &  <0.001\textbf{***} &    -8.192387 & -0.560020 &  214 \\
        anger   &  <0.001\textbf{***} &    -6.771028 & -0.462858 &  0.052 &    -1.945685 & -0.133004 &  214 \\
        \bottomrule
        \end{tabular}
    }
    \caption{The statistics for the cosine distance of GANonymization the DeepPrivacy2, and CIAGAN method to the original based on dataset a), b), and c). If a p-value is less than 0.05, it is flagged with one star (*). If a p-value is less than 0.01, it is flagged with 2 stars (**). If a p-value is less than 0.001, it is flagged with three stars (***)}
    \label{tab:emotion_pvalues}
\end{table}

\paragraph{Training Scenario Evaluation.}
The full evaluation results for the training scenario can be found in Table~\ref{table:classification_report_emotion} in the appendix, whereas the confusion matrices for the single models are shown in Figure~\ref{fig:emotion_classifier_conf_matrix} in the appendix.
For the AffectNet dataset, the classifier trained on the original data achieved an overall F1 score of $0.58$.
In contrast, the classifier trained on the data anonymized with GANonymization achieved an overall F1 score of $0.37$.
DeepPrivacy2 led to a worse performance, reaching only an F1 score of $0.30$.
CIAGAN could acquire a slightly increased F1 score of $0.38$ than our method.

The other datasets continue the trend for DeepPrivacy2 but worsen the performance for CIAGAN: CK+ (Original data: $0.99$, GANonymization data: $0.69$, DeepPrivacy2 data: $0.46$, CIAGAN data: $0.62$) and FACES (Original data: $0.97$, GANonymization data: $0.81$, DeepPrivacy2 data: $0.67$, CIAGAN data: $0.75$).

\subsubsection{Discussion}

\paragraph{Inference Scenario Evaluation.}
The overall results indicate the superior performance of our approach in preserving facial expressions.

It outperformed the mean distance of DeepPrivacy2 for all emotions except \emph{Fear} and \emph{Happy} in AffectNet.
However, we did not find statistical evidence (see Table~\ref{tab:emotion_pvalues}) for the performance differences for \emph{all} of those classes in CK+ and FACES (which might be because those two datasets include a substantially lower amount of images than AffectNet).
This could be because many predictions from \emph{Fear} and \emph{Happy} of the synthesized images of our approach were mixed classified (see Figure~\ref{fig:emotion_classifier_conf_matrix} in the appendix).
For example, the emotions \emph{Happy} and \emph{Surprise} were mainly predicted as \emph{Fear} by our classification model.

Compared to the synthesized images by CIAGAN, the cosine distances are closer to our method.
CIAGAN preserved \emph{Fear}, \emph{Happy}, and \emph{Sadness} significantly better in the AffectNet dataset (see Table~\ref{tab:emotion_pvalues}).
Additionally, the emotions \emph{Contempt}, \emph{Fear}, and \emph{Surprise} also performed significantly better in CK+ judging by the mean distance.
In the FACES dataset, CIAGAN outperformed our method only for the emotion \emph{Happy}.
However, the results from the significance test for the FACES dataset in Table~\ref{tab:emotion_pvalues} show that it does not have any statistical significance.
An explanation for the small gap between the cosine distances from our method and CIAGAN could be a similar approach to the facial landmarks.
The facial landmarks preserve the facial expression mostly accurately.
However, in increasing the number of facial landmark points with our approach, it becomes clear that the affective state preservation can be enhanced.


\paragraph{Training Scenario Evaluation.}
Here, we could observe that data obtained through the anonymization methods led to substantially worse F1 scores for the trained classifiers than the original data.
However, GANonymization still performed better for each dataset, except for a very slightly worsening performance in the AffectNet dataset compared to CIAGAN.

\subsection{Analysis of Facial Feature Anonymization}
\label{sec:analysis}
To better understand which features are being preserved and which are discarded by GANonymization, we performed an analysis using a pre-trained model for facial feature classification on the CelebA \cite{liu2015faceattributes} dataset. 
By analyzing how the predictions of that model change when applied to original versus anonymized images, we aim to infer insights about which facial features our model removes.

\subsubsection{Dataset}
We've chosen the CelebA dataset due to its vast amount of 202,599 face images with 10,177 identities and 40 binary features representing about facial attributes per subject.
For example, those attributes entail eyeglasses, hairstyle, hair color, facial shape, and beard.
Thus, this dataset is well suited for analyzing which attributes might change with our anonymization method. 
However, it should be noted that the dataset contains primarily images of young celebrities - as those might visually not be a representative sample of the entirety of people, it might influence the analysis. 
We applied our pre-processing pipeline with face extraction and face segmentation on the dataset and received a training and validation set of 166,223 and 20,259 images, respectively.

\subsubsection{Setup}
Similar to section~\ref{subsec:anonymization_performance} and section~\ref{subsec:preserved_facial_expressions}, the analysis of which traits of the original face images are removed through our anonymization pipeline is based on utilizing an auxiliary classifier to compare original versus anonymized images.
We trained the same model architecture described in Section~\ref{subsubsec:emotion_setup}, but this time to classify facial features rather than emotions.
The only changes made to the architecture were matching the output layer to fit the number of features incorporated in the CelebA dataset, switching to a binary-cross-entropy loss, and changing the output activation function to Sigmoid, as in this case, we dealt with a multi-label task (i.e., multiple traits can be present at once).
Here, each feature can be interpreted as a facial trait that is apparent in the model's face input.
Exemplary anonymized images for CelebA can be seen in Figure~\ref{fig:celeba_synthesized_samples}.
The performance of the classification model can be looked up in the appendix in Figure~\ref{fig:classifier_conf_matrix_face_attribute} and Table~\ref{table:classification_report_face_attributes}.

\begin{figure}[!ht]
    \centering
    \includegraphics[width=1\textwidth]{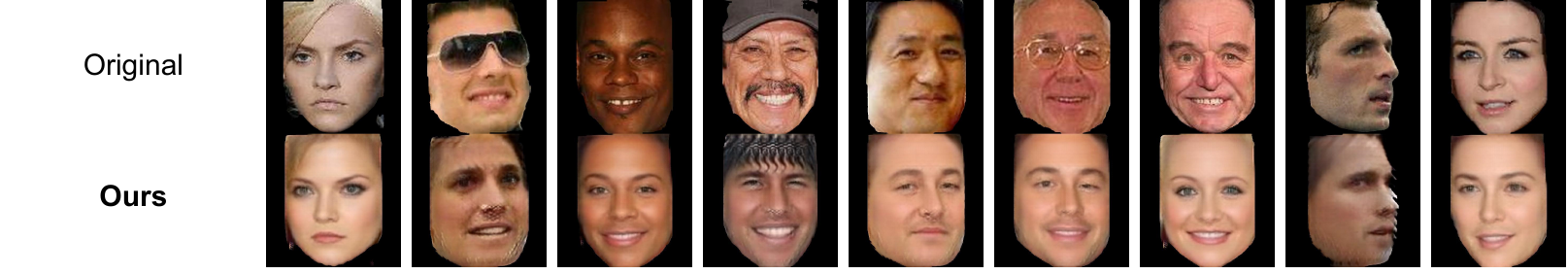}
    \caption{Sample of synthesized faces based on the CelebA dataset.}
    \label{fig:celeba_synthesized_samples}
\end{figure}

\subsubsection{Metric}
To examine which of those traits get removed, for each trait we take the subset of images in the original dataset where the classifier predicted that trait, i.e., the classifier assigned it a probability of $>0.5$.
Subsequently, we assess the portion of anonymized versions of those images where the classifier did not predict the respective trait.

\subsubsection{Results}
\begin{figure}[!ht]
    \centering
    \includegraphics[width=\columnwidth]{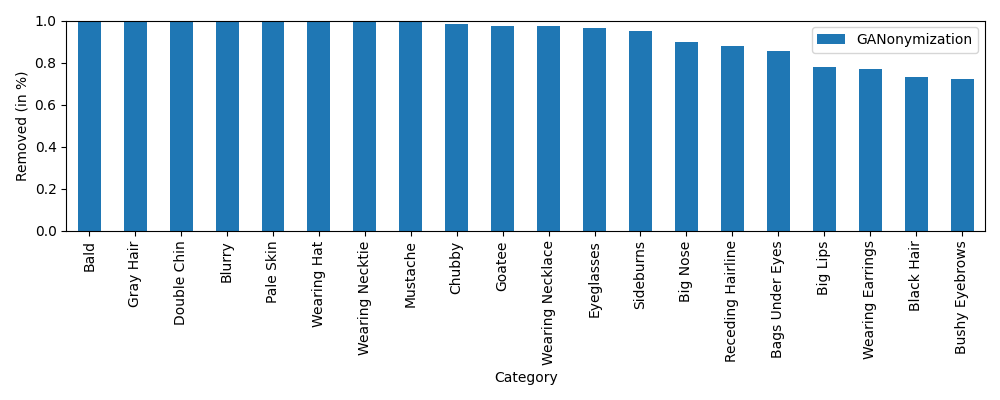}
    \includegraphics[width=\columnwidth]{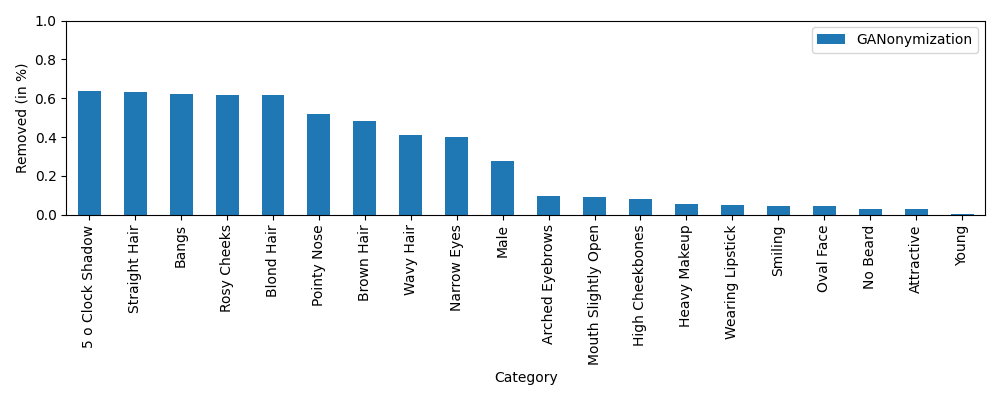}
    \caption{The removed categories in \% over the total number of available samples for each category in the CelebA dataset.}
    \label{fig:celeba_classification}
\end{figure}

The results are depicted in Figure~\ref{fig:celeba_classification}. Here, we ordered the features according to the percentage of cases where they have been removed. 
The actual percentages are in the appendix in Table~\ref{tab:celeba_classification}.

\subsubsection{Discussion}
As can be seen in the results, some traits were removed in 100\% of the cases, whilst others were preserved in almost all images.
Traits that refer to head or facial hair, e.g., \emph{Bald}, \emph{Gray Hair}, \emph{Mustache}, or \emph{Goatee} are removed quite frequently. 
This is not surprising since the only information that our re-synthesis model can rely on is the landmark representation of the input face.
Also, wearing specific accessories like neckties, hats, or necklaces is not encoded in landmark representations, resulting in them getting reliably removed.
The \emph{Smiling} feature, which is highly correlated to emotional expressions, gets preserved quite well, which again supports our claim of being able to preserve such expressions.
On the other hand, a surprising observation is that \emph{Heavy Makeup} and \emph{Wearing Lipstick} predominantly are getting preserved. 
The training data we used for our GAN model is a possible explanation.
For that, the CelebA dataset, containing exclusively celebrity face images, was used, too.
In the world of celebrities, it is common practice for women to dress up and apply makeup for their appearance at public events.
As the GAN model aims to resemble the data distribution imposed by the training data, these traits are also apparent in the anonymized versions. 
The same goes for features like \emph{No Beard} or \emph{Young} - the vast majority of subjects in the CelebA dataset are relatively young and do not wear a beard \cite{rudd2016moon}.
Besides that, an interesting observation is that the \emph{Chubby} trait was removed in the vast majority of cases where it was apparent.
Intuitively, the facial landmark representation should have covered that trait, but apparently, it wasn't.
The same goes for \emph{Big Nose} and \emph{Big Lips} - which were also removed frequently. 
Removing those traits advocates our approach since they are typical examples of features that could introduce unfairness and bias into datasets.

The feature \emph{Male} got removed in 27.62\% of the cases. 
It has to be noted that there is no \emph{Female} feature in the dataset, and as such, the absence of the \emph{Male} trait is mainly interpreted as the face being female.  
Therefore, it is good that the \emph{Male} trait did not get removed in 100\% of the cases - which would mean that the anonymized versions are \emph{always} female.
Here, a medium removal rate indicates that the gender sometimes changes and sometimes does not, indicating that it indeed gets diluted by GANonymization. 

Finally, the feature \emph{Blurry} was removed in over 99\% of the cases. 
Although this trait doesn't refer to the face itself but to the image quality, it is a good indicator that the results of GANonymization are of high quality - even if the original images are not.

\section{Conclusion}
This research aimed to evaluate the anonymization performance using our method - GANonymization.
Our method is a generative adversarial network (GAN) based image-to-image translation approach to anonymize faces and preserve their original facial expression.
Facial landmarks serve as image input into a \textit{pix2pix} architecture to re-synthesize high-quality, anonymized versions of the input face image.

First, we measured the efficiency of our approach in removing identifiable facial attributes to increase the anonymity of the given individual face.
Our method proved its anonymization performance in the chosen metric on the WIDER dataset.

Second, we evaluated the performance regarding preserving emotional facial expressions on the AffectNet, CK+, and FACES datasets.
Our approach significantly outperformed DeepPrivacy2 in most categories. 
However, DeepPrivacy2 significantly outperformed our approach in the emotion \emph{Fear} and \emph{Happy} from the AffectNet dataset.
Compared to CIAGAN we could show a significant improvement in most of the preserved emotional facial expressions for \emph{Neutral}, \emph{Anger}, \emph{Contempt} (in AffectNet), \emph{Disgust}, \emph{Fear} (in FACES), \emph{Happy} (in CK+), \emph{Sadness} (in CK+ and FACES), and \emph{Surprise} (in AffectNet).
Furthermore, a noticeable quality difference in the image could be seen between the different methods.
Here, our method showed the highest quality in the synthesized faces.

Last, analyzing facial traits removed by our approach showed that some traits were eliminated in almost 100\% of the cases while others were preserved.
Especially jewelry, clothing, and hair, e.g., \emph{Bald}, \emph{Gray Hair}, \emph{Mustache}, or \emph{Goatee} are removed quite reliably.

In future efforts, training the GAN with a wider variety of facial expressions and facial traits might be supportive in increasing the overall performance in preserving the facial expressions, especially in adding more diversity to the generated faces.
Finally, it has to be studied how GANonymization can be transferred to other, not emotion-related contexts. 
Therefore, suiting intermediate representations for the specific tasks have to be investigated. 
The medical domain is one field where adapting our approach might have a  major impact.
Here, anonymizing patient photos could reduce the sparsity of available data, which is crucial for that field.
Doing so might enhance the data basis researchers can work with without being restricted by data privacy regulations.


\bibliographystyle{ACM-Reference-Format}
\bibliography{main}
\appendix

\section{Appendix}

\begin{figure}[H]
    \centering
    \setlength{\tabcolsep}{0pt}
    \begin{tabular}{cccc}
        \multicolumn{4}{c}{\textbf{AffectNet}} \\
        Original & GANonymization & DeepPrivacy2 & CIAGAN \\
        \includegraphics[width=0.25\columnwidth]{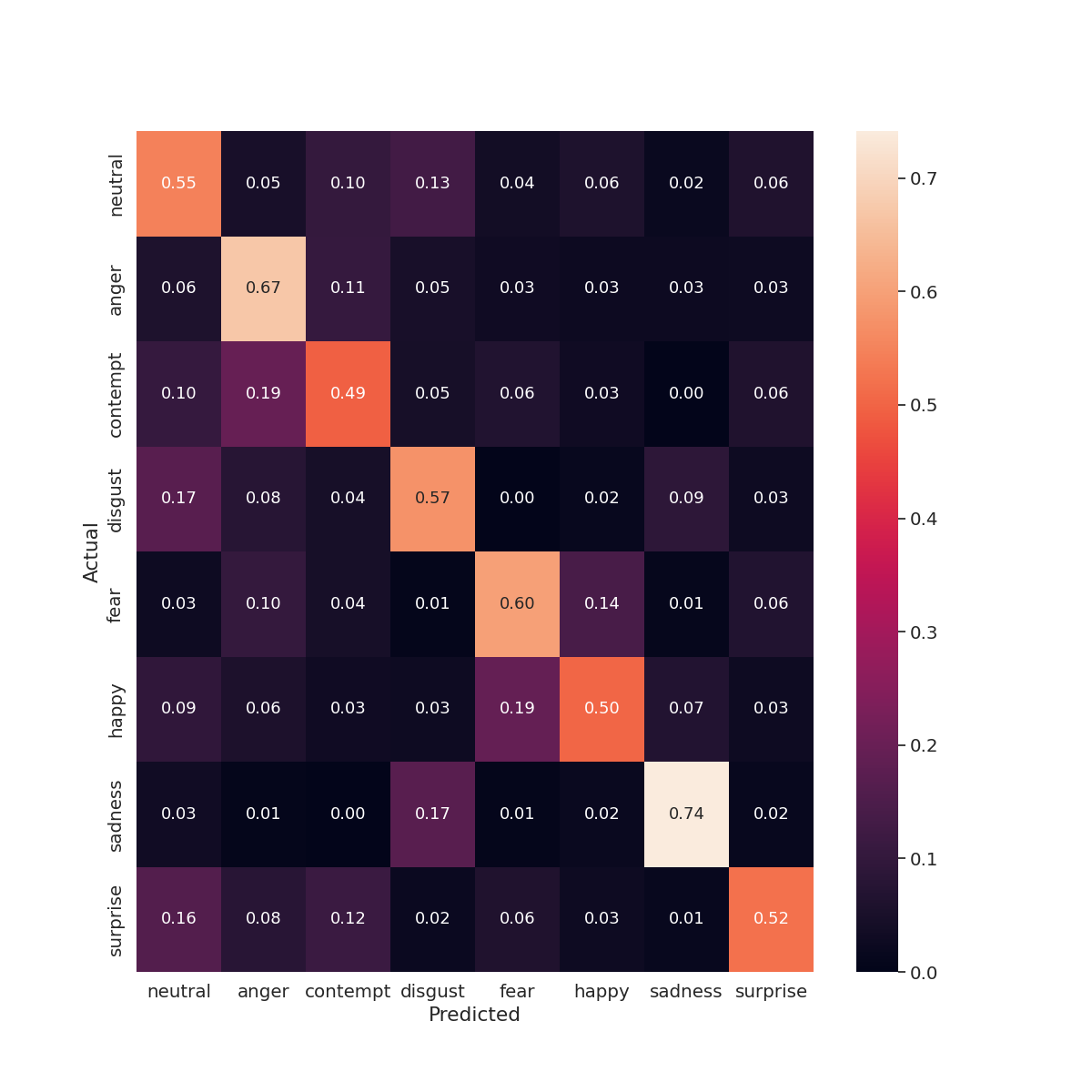} &   \includegraphics[width=0.25\columnwidth]{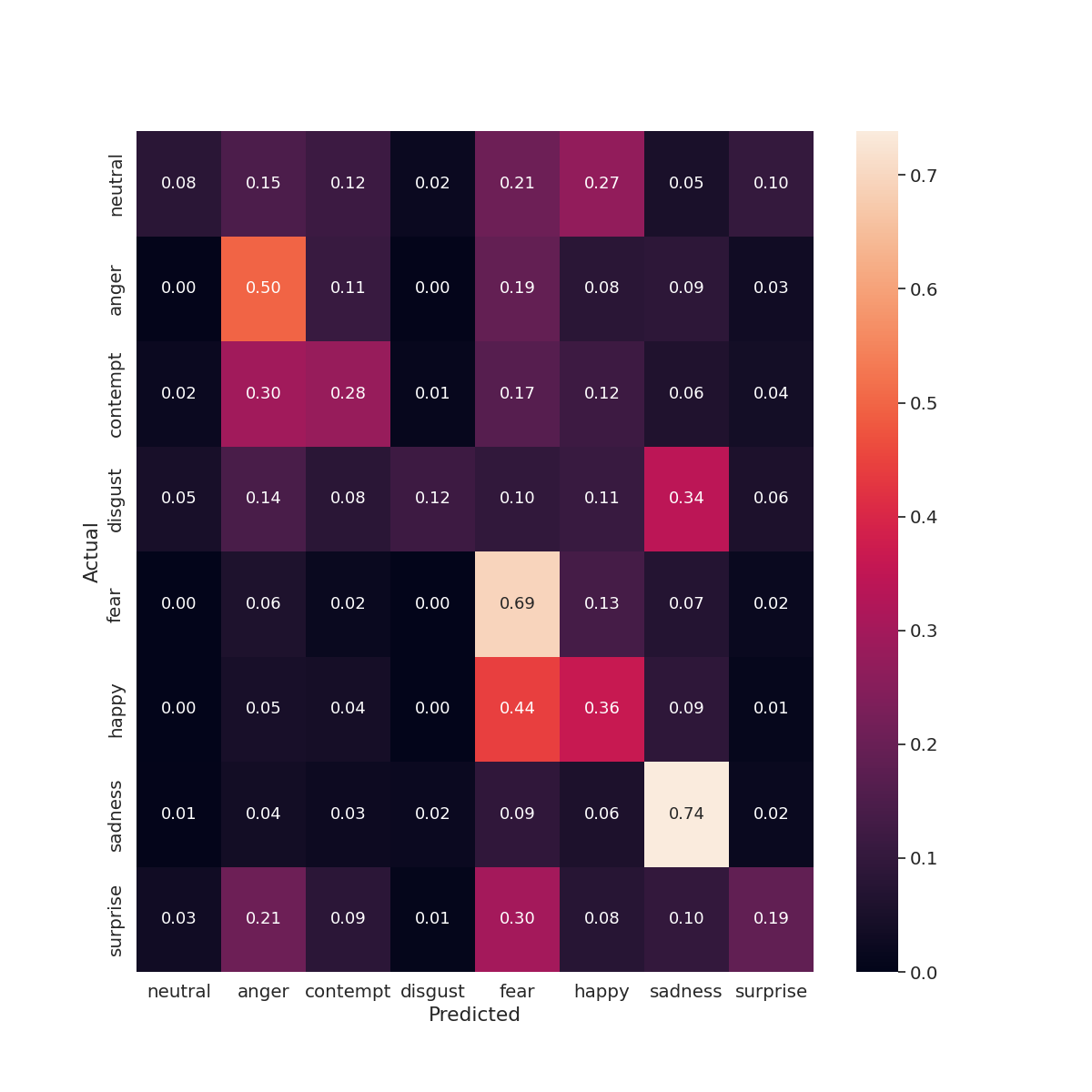} &   \includegraphics[width=0.25\columnwidth]{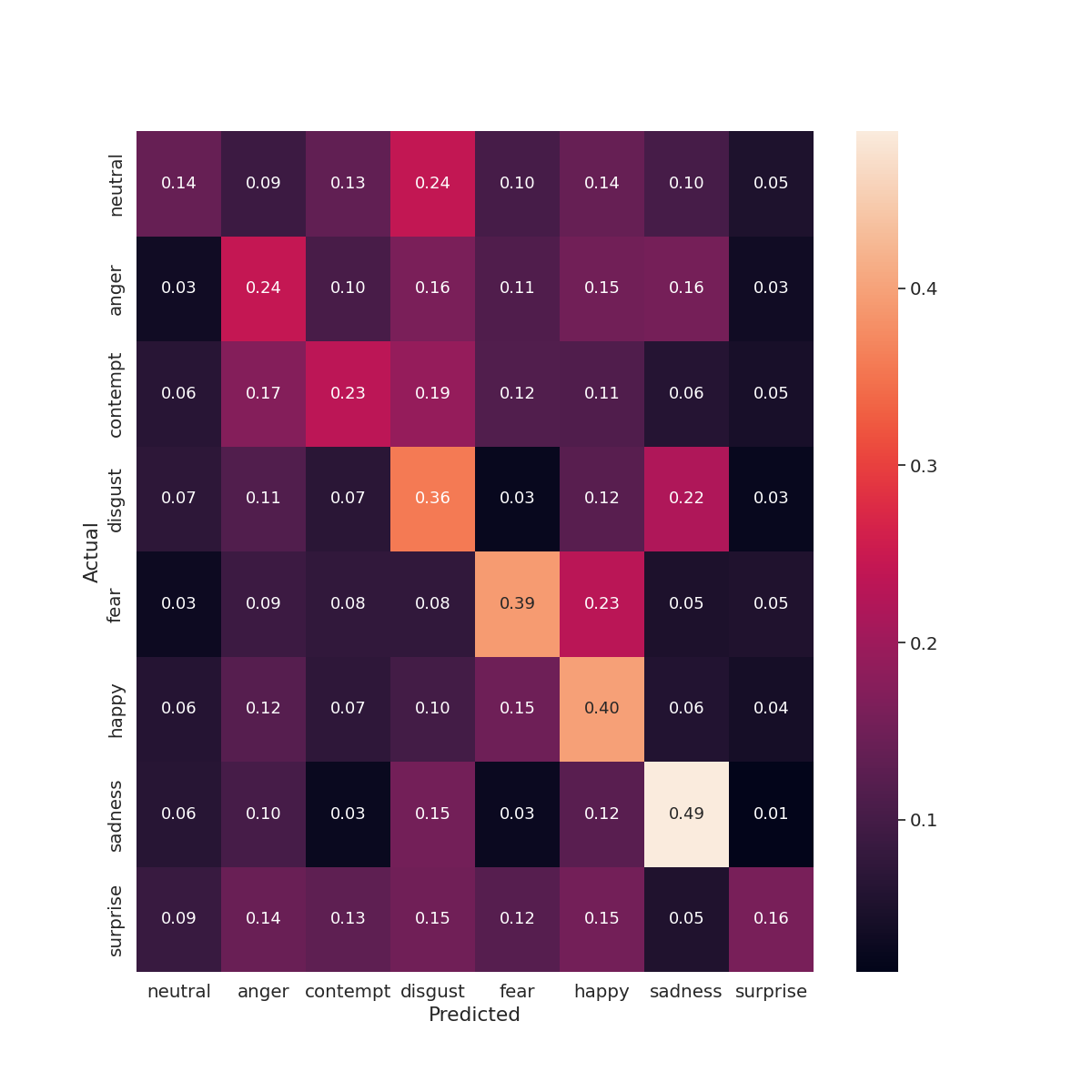} &   \includegraphics[width=0.25\columnwidth]{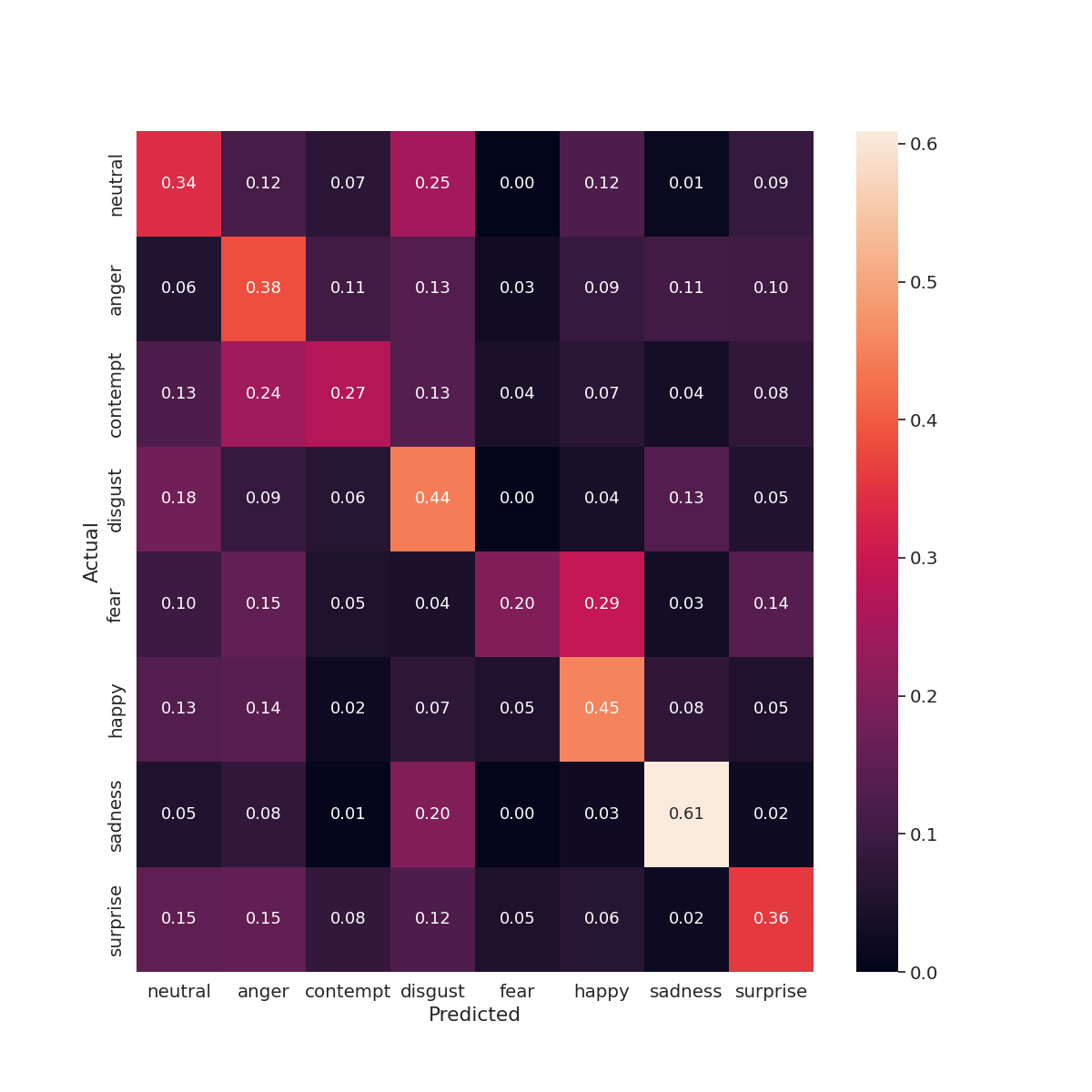} \\
        \hline
        \multicolumn{4}{c}{\textbf{CK+}} \\
        Original & GANonymization & DeepPrivacy2 & CIAGAN \\
        \includegraphics[width=0.25\columnwidth]{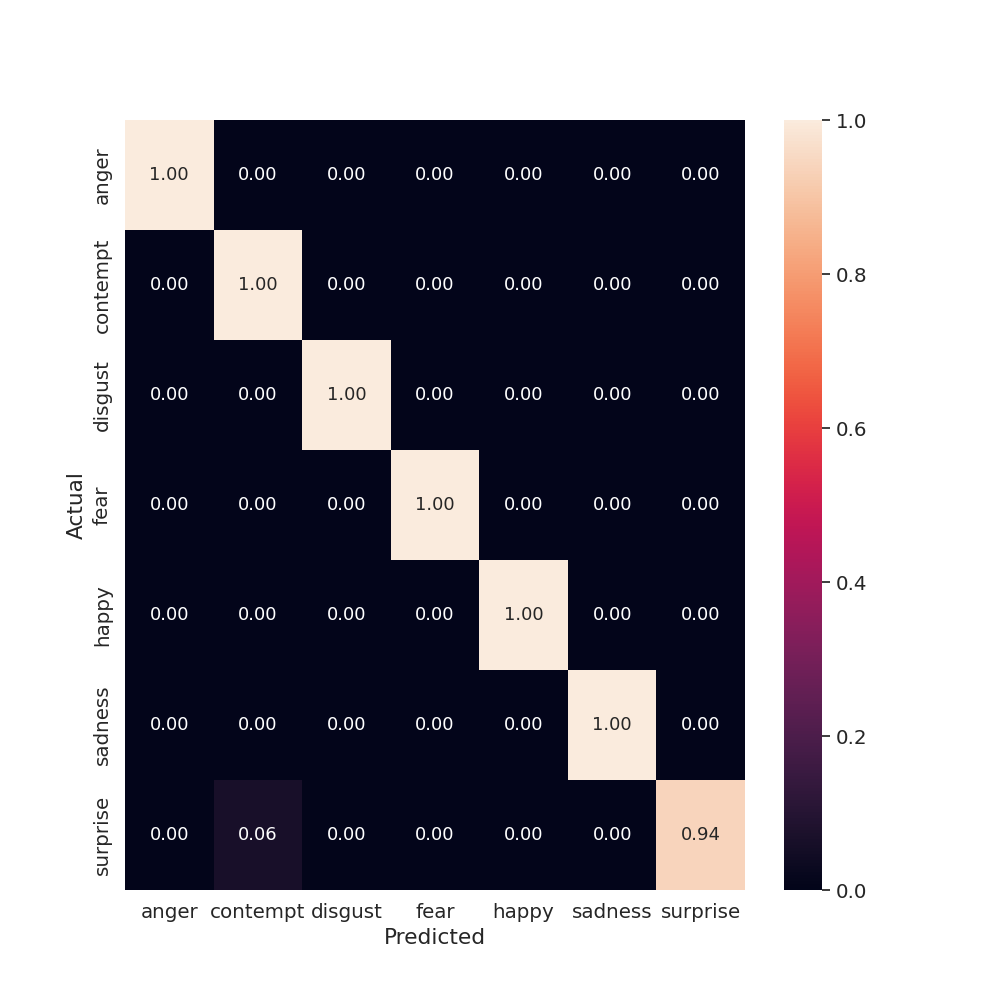} &   \includegraphics[width=0.25\columnwidth]{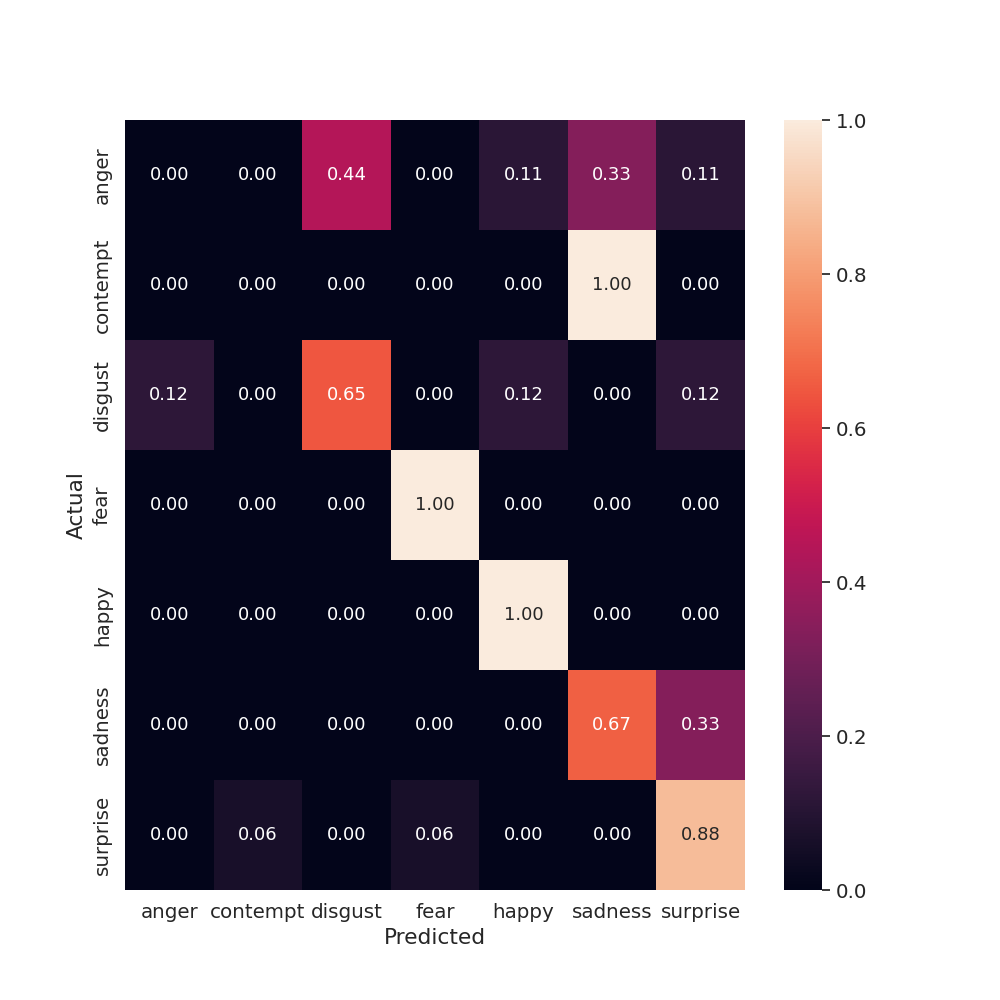} &   \includegraphics[width=0.25\columnwidth]{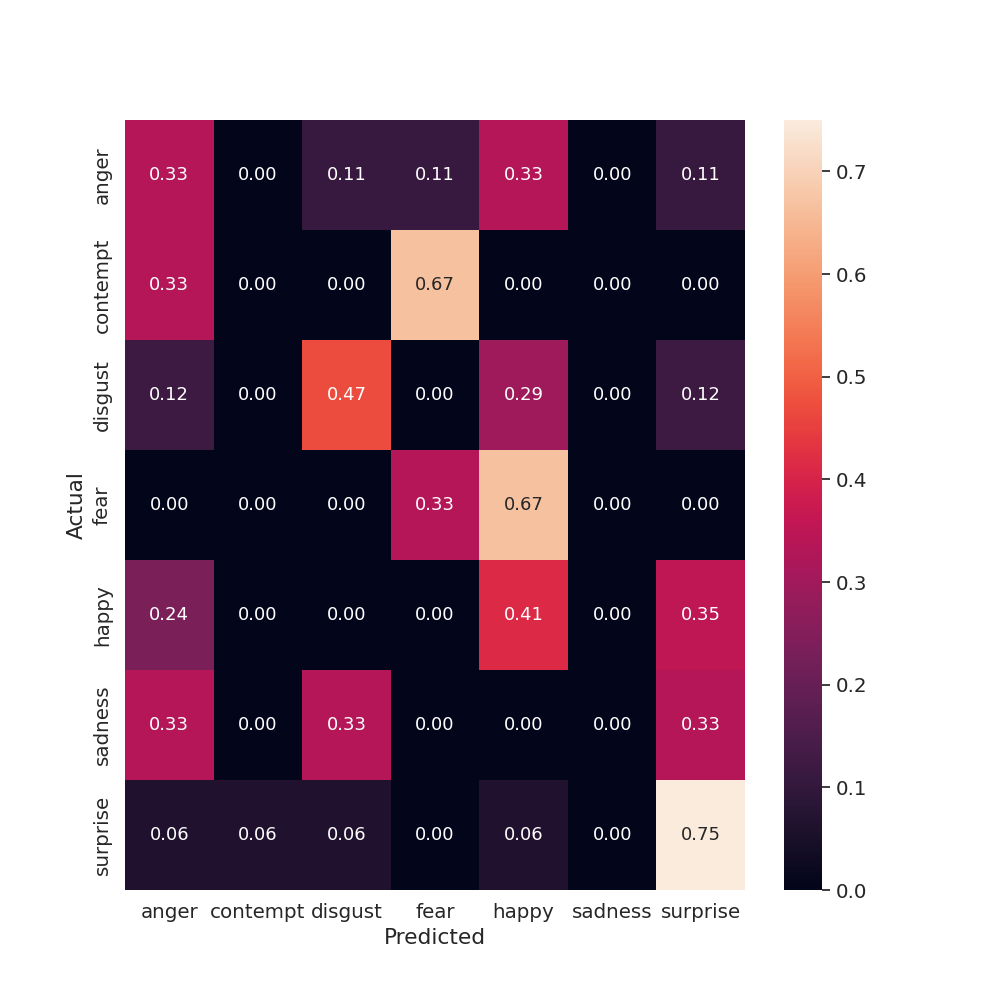} &         \includegraphics[width=0.25\columnwidth]{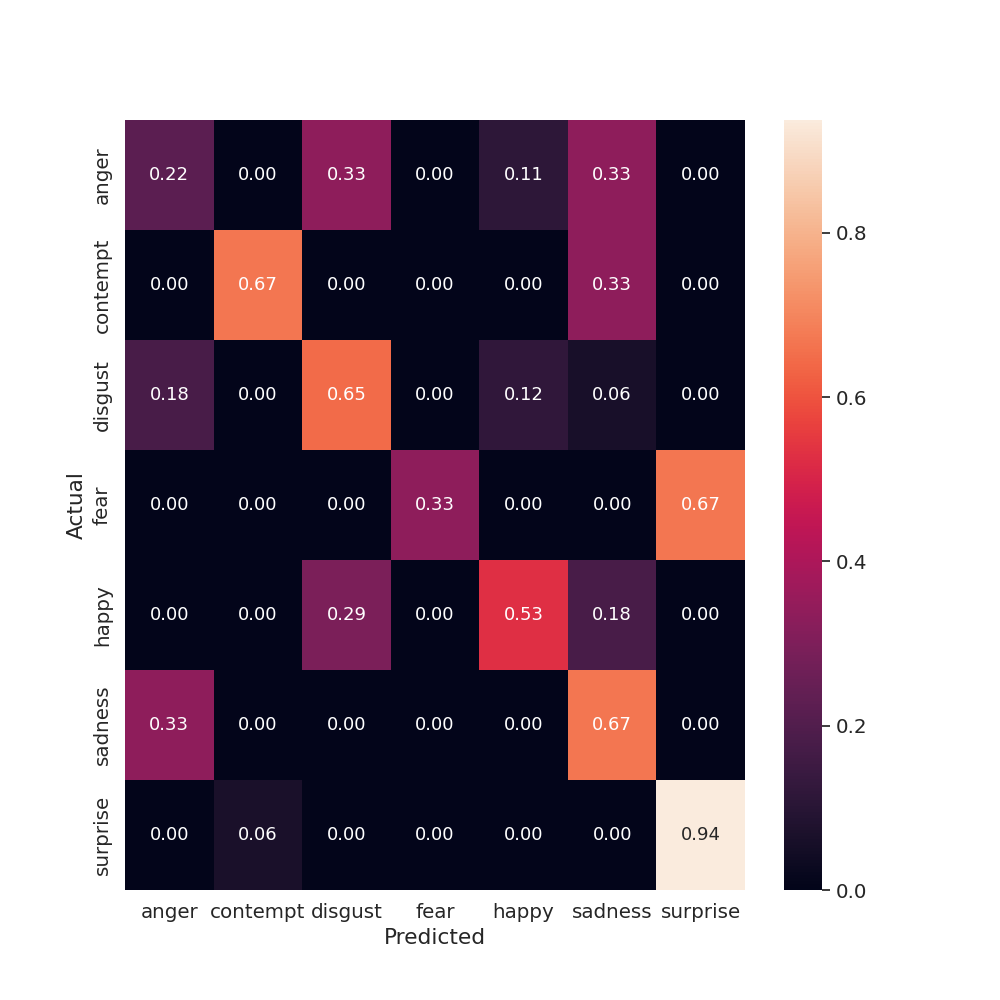}\\
        \hline
        \multicolumn{4}{c}{\textbf{FACES}} \\
        Original & GANonymization & DeepPrivacy2 & CIAGAN \\
        \includegraphics[width=0.25\columnwidth]{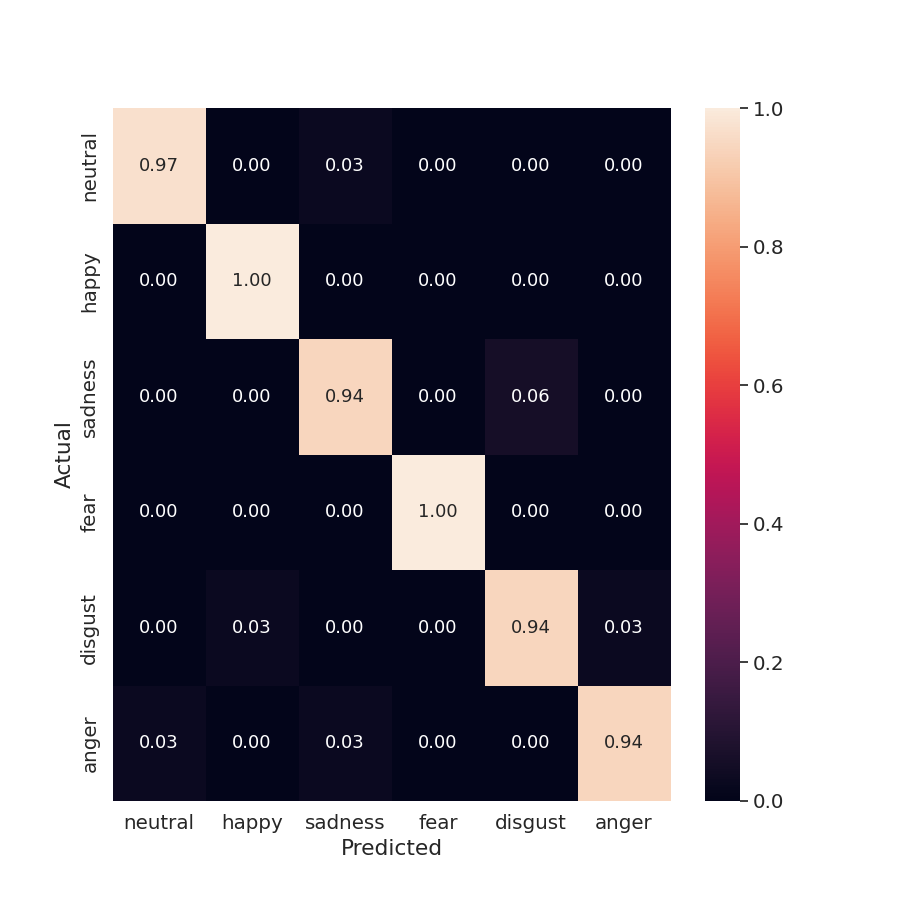} &   \includegraphics[width=0.25\columnwidth]{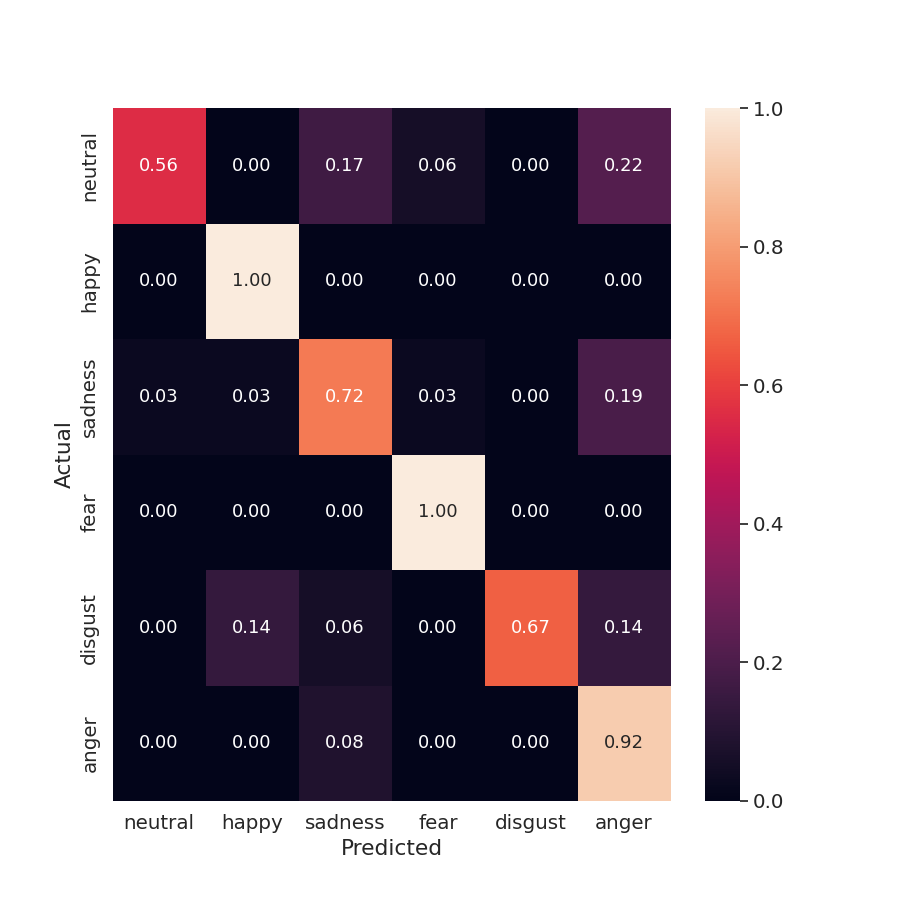} &   \includegraphics[width=0.25\columnwidth]{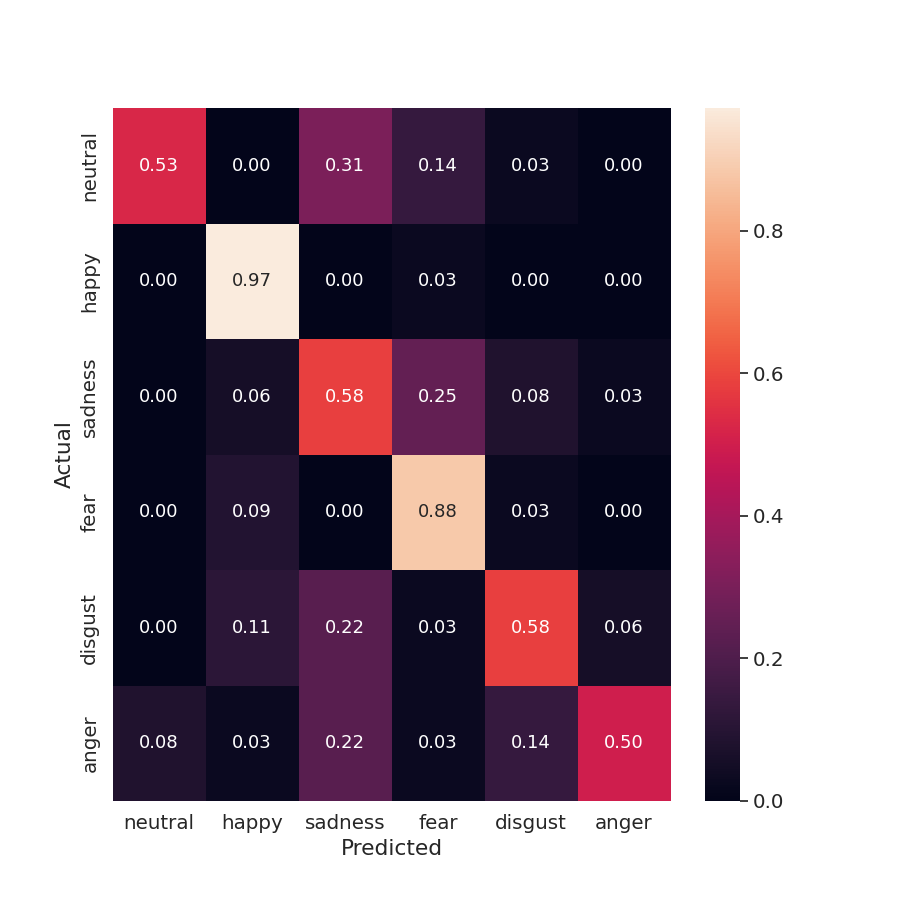}  &   \includegraphics[width=0.25\columnwidth]{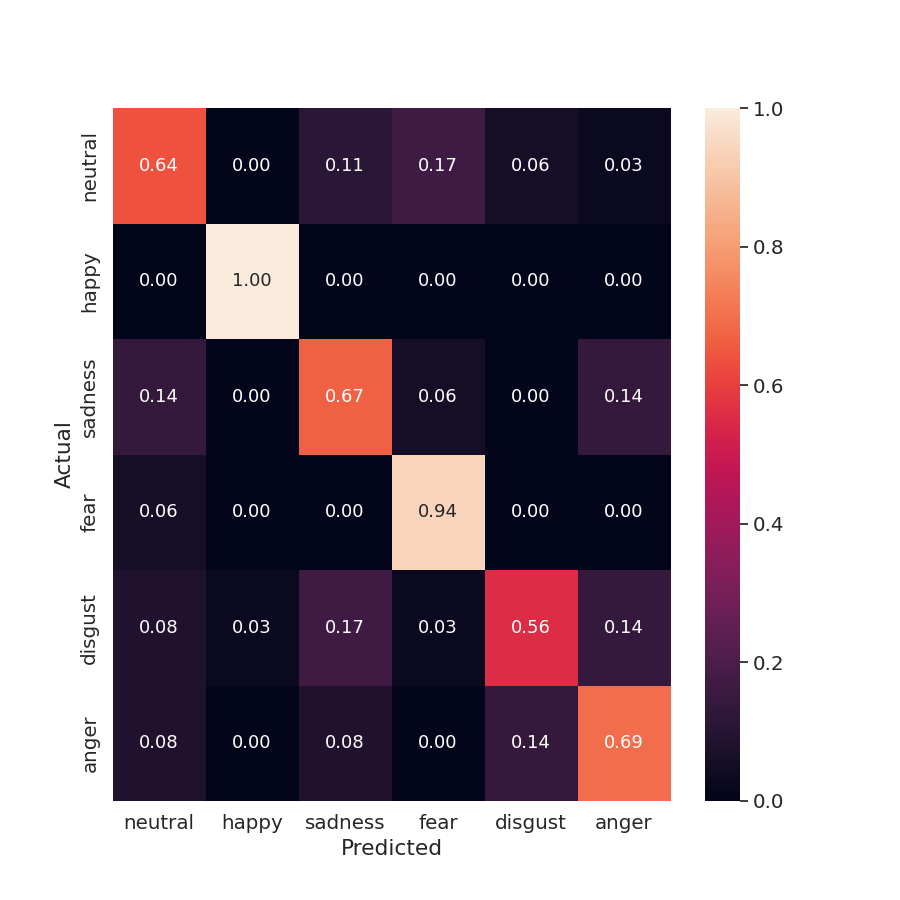} \\
    \end{tabular}
    \caption{For each specified dataset a multi-class classification model was trained. Accordingly, the confusion matrices depict the classification model's performance on the validation sets. The column "Original" model was trained on the original images from the training split. The column "GANonymization" and "DeepPrivacy2" contains the models trained on the synthesized images of the training split, respectively.}
    \label{fig:emotion_classifier_conf_matrix}
\end{figure}

\begin{table}[H]
    \centering
    \caption{For each specified dataset a multi-class classification model was trained. Accordingly, the classification reports show the classification model's performance on the validation sets. The column "Original" model was trained on the original images from the training split. The column "GANonymization" and "DeepPrivacy2" contains the models trained on the synthesized images of the training split, respectively. (P) Precision; (R) Recall; (F1) F1-Score; (N) Support}
    \renewcommand{\arraystretch}{0.9} 
    \setlength{\tabcolsep}{3pt}
    \begin{tabular}{|c | l | c c c | c c c | c c c | c c c | c |} 
         \hline
          & & \multicolumn{3}{|c|}{Original} & \multicolumn{3}{|c|}{GANonymization} & \multicolumn{3}{|c|}{DeepPrivacy2} & \multicolumn{3}{|c|}{CIAGAN} & \\
         Dataset & & P & R & F1 & P & R & F1 & P & R & F1 & P & R & F1 & N \\
         \hline\hline
         \multirow{11}{*}{AffectNet} & Neutral & 0.49 & 0.41 & 0.45 & 0.43 & 0.08 & 0.14 & 0.26 & 0.14 & 0.18 & 0.30 & 0.34 & 0.32 & 360 \\
         & Anger & 0.58 & 0.59 & 0.58 & 0.34 & 0.50 & 0.41 & 0.22 & 0.24 & 0.23 & 0.28 & 0.38 & 0.32 & 346 \\
         & Contempt & 0.48 & 0.67 & 0.56 & 0.37 & 0.28 & 0.32 & 0.28 & 0.23 & 0.25 & 0.41 & 0.27 & 0.33 & 354 \\
         & Disgust & 0.58 & 0.54 & 0.56 & 0.62 & 0.12 & 0.20 & 0.25 & 0.36 & 0.29 & 0.32 & 0.44 & 0.38 & 357 \\
         & Fear & 0.65 & 0.62 & 0.63 & 0.32 & 0.69 & 0.44 & 0.38 & 0.39 & 0.38 & 0.54 & 0.20 & 0.29 & 357 \\
         & Happy & 0.57 & 0.51 & 0.53 & 0.30 & 0.36 & 0.33 & 0.28 & 0.40 & 0.33 & 0.39 & 0.45 & 0.42 & 362 \\
         & Sadness & 0.68 & 0.77 & 0.72 & 0.48 & 0.74 & 0.58 & 0.41 & 0.49 & 0.45 & 0.60 & 0.61 & 0.60 & 352 \\
         & Surprise & 0.66 & 0.55 & 0.60 & 0.39 & 0.19 & 0.25 & 0.36 & 0.16 & 0.22 & 0.39 & 0.36 & 0.37 & 337 \\
         \cline{2-15}
         & accuracy &  &  & 0.58 &  &  & 0.37 &  &  & 0.30 & & & 0.38 & 2825 \\
         & macro avg & 0.59 & 0.58 & 0.58 & 0.41 & 0.37 & 0.33 & 0.31 & 0.30 & 0.29 & 0.40 & 0.38 & 0.38 & 2825 \\
         & weighted avg & 0.58 & 0.58 & 0.58 & 0.41 & 0.37 & 0.33 & 0.30 & 0.30 & 0.29 & 0.40 & 0.38 & 0.38 & 2825 \\
         \hline\hline
         \multirow{10}{*}{CK+}& Anger & 1.00 & 1.00 & 1.00 & 0.00 & 0.00 & 0.00 & 0.25 & 0.33 & 0.29 & 0.33 & 0.22 & 0.27 & 9 \\
         & Contempt & 0.75 & 1.00 & 0.86 & 0.00 & 0.00 & 0.00 & 0.00 & 0.00 & 0.00 & 0.67 & 0.67 & 0.67 & 3 \\
         & Disgust & 1.00 & 1.00 & 1.00 & 0.73 & 0.65 & 0.69 & 0.73 & 0.47 & 0.57 & 0.58 & 0.65 & 0.61 & 17 \\
         & Fear & 1.00 & 1.00 & 1.00 & 0.75 & 1.00 & 0.86 & 0.25 & 0.33 & 0.29 & 1.00 & 0.33 & 0.50 & 3 \\
         & Happy & 1.00 & 1.00 & 1.00 & 0.85 & 1.00 & 0.92 & 0.39 & 0.41 & 0.40 & 0.75 & 0.53 & 0.62 & 17 \\
         & Sadness & 1.00 & 1.00 & 1.00 & 0.25 & 0.67 & 0.36 & 0.00 & 0.00 & 0.00 & 0.20 & 0.67 & 0.31 & 3 \\
         & Surprise & 1.00 & 0.94 & 0.97 & 0.78 & 0.88 & 0.82 & 0.55 & 0.75 & 0.63 & 0.88 & 0.94 & 0.91 & 16 \\
         \cline{2-15}
         & accuracy &  &  & 0.99 &  &  & 0.69 &  &  & 0.46 & & & 0.62 & 68 \\
         & macro avg & 0.96 & 0.99 & 0.97 & 0.48 & 0.60 & 0.52 & 0.31 & 0.33 & 0.31 & 0.63 & 0.57 & 0.55 & 68 \\
         & weighted avg & 0.99 & 0.99 & 0.99 & 0.62 & 0.69 & 0.65 & 0.45 & 0.46 & 0.44 & 0.67 & 0.62 & 0.62 & 68 \\
         \hline\hline
         \multirow{9}{*}{FACES} & neutral & 0.97 & 0.97 & 0.97 & 0.95 & 0.56 & 0.70 & 0.86 & 0.53 & 0.66 & 0.64 & 0.64 & 0.64 & 36 \\
         & happy & 0.97 & 1.00 & 0.99 & 0.86 & 1.00 & 0.92 & 0.78 & 0.97 & 0.86 & 0.97 & 1.00 & 0.99 & 36 \\
         & sadness & 0.94 & 0.94 & 0.94 & 0.70 & 0.72 & 0.71 & 0.44 & 0.58 & 0.50 & 0.65 & 0.67 & 0.66 & 36 \\
         & fear & 1.00 & 1.00 & 1.00 & 0.92 & 1.00 & 0.96 & 0.64 & 0.88 & 0.74 & 0.78 & 0.94 & 0.85 & 34 \\
         & disgust & 0.94 & 0.94 & 0.94 & 1.00 & 0.67 & 0.80 & 0.68 & 0.58 & 0.63 & 0.74 & 0.56 & 0.63 & 36 \\
         & anger & 0.97 & 0.94 & 0.96 & 0.62 & 0.92 & 0.74 & 0.86 & 0.50 & 0.63 & 0.69 & 0.69 & 0.69 & 36 \\
         \cline{2-15}
         & accuracy &  &  & 0.97 &  &  & 0.81 &  &  & 0.67 & & & 0.75 & 214 \\
         & macro avg & 0.97 & 0.97 & 0.97 & 0.84 & 0.81 & 0.81 & 0.71 & 0.67 & 0.67 & 0.75 & 0.75 & 0.74 & 214 \\
         & weighted avg & 0.97 & 0.97 & 0.97 & 0.84 & 0.81 & 0.80 & 0.71 & 0.67 & 0.67 & 0.75 & 0.75 & 0.74 & 214 \\
         \hline
    \end{tabular}
    \label{table:classification_report_emotion}
\end{table}

\begin{figure}[H]
    \centering
    \renewcommand{\arraystretch}{0.1} 
    \begin{tabular}{c}
      Original \\
      \includegraphics[width=0.75\columnwidth]{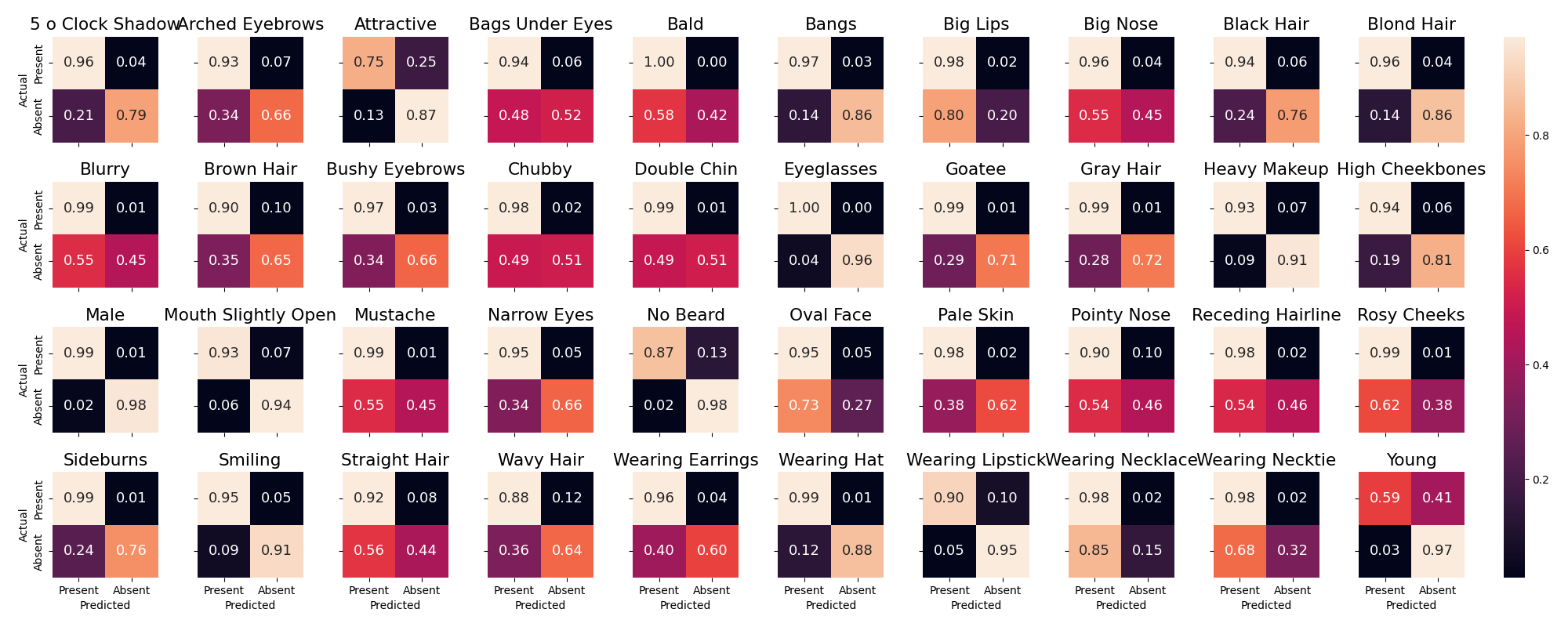} \\
      GANonymization \\
      \includegraphics[width=0.75\columnwidth]{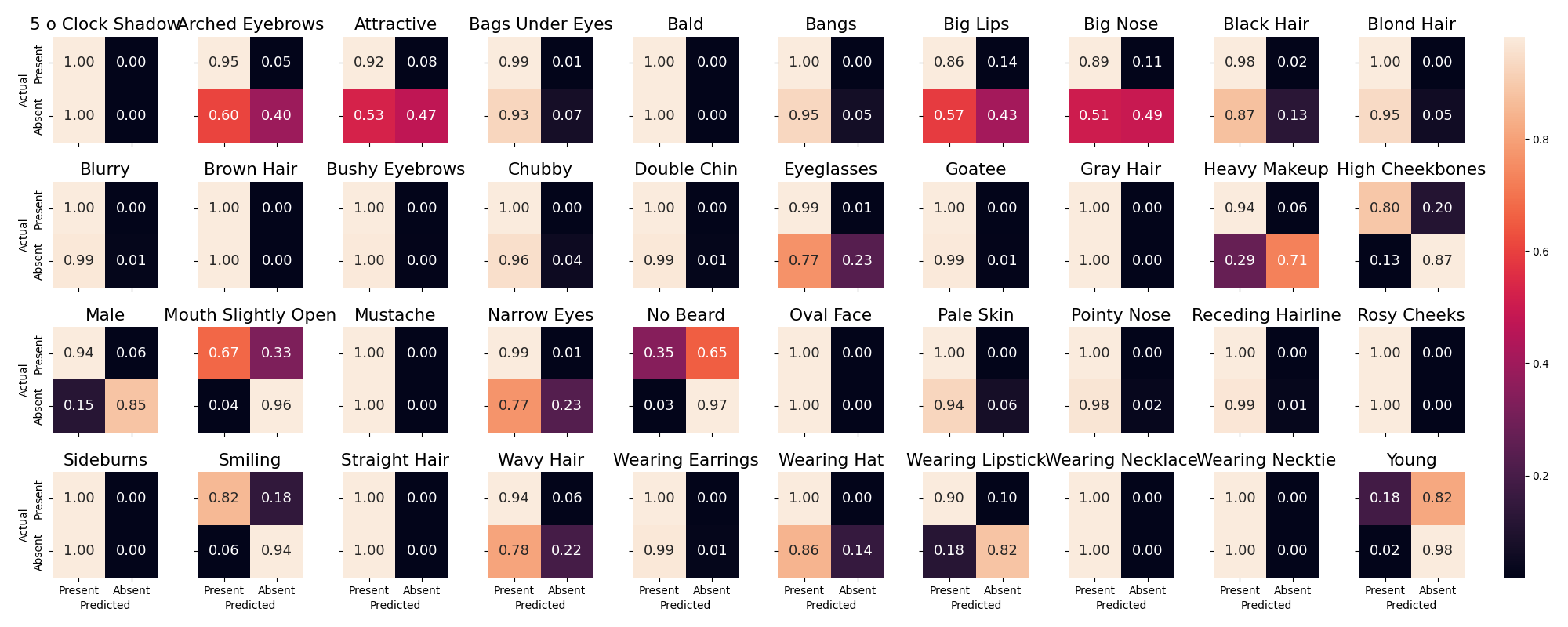} \\
      DeepPrivacy2 \\
      \includegraphics[width=0.75\columnwidth]{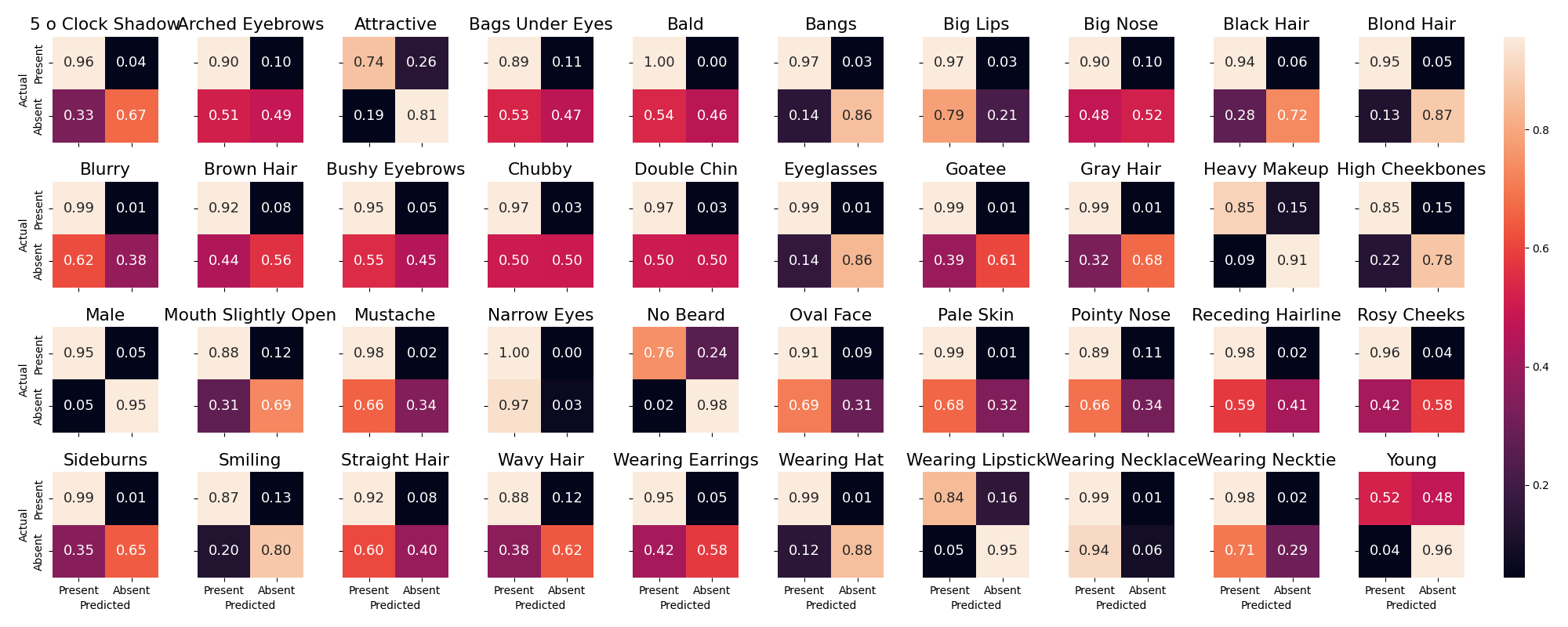} \\
      CIAGAN \\
      \includegraphics[width=0.75\columnwidth]{figures/deepprivacy2/celeba_confusion_matrix.png} \\
    \end{tabular}
    \caption{A multi-label classification model was trained on the CelebA dataset. Accordingly, the confusion matrices depict the classification model's performance on the validation sets. The column "Original" model was trained on the original images from the training split. The column "GANonymization" and "DeepPrivacy2" contains the models trained on the synthesized images of the training split, respectively.}
    \label{fig:classifier_conf_matrix_face_attribute}
\end{figure}

\begin{table}[H]
    \centering
    \caption{A multi-label classification model was trained on the CelebA dataset. Accordingly, the classification reports show the classification model's performance on the validation sets for each label. The column "Original" model was trained on the original images from the training split. The column "GANonymization" and "DeepPrivacy2" contains the models trained on the synthesized images of the training split, respectively. (P) Precision; (R) Recall; (F1) F1-Score; (N) Support}
    \renewcommand{\arraystretch}{0.8} 
    \setlength{\tabcolsep}{3pt}
    \begin{tabular}{|l|ccc|ccc|ccc|ccc|c|} 
        \hline
         & \multicolumn{3}{|c|}{Original} & \multicolumn{3}{|c|}{GANonymization} & \multicolumn{3}{|c|}{DeepPrivacy2} & \multicolumn{3}{|c|}{CIAGAN} &  \\
         & P & R & F1 & P & R & F1 & P & R & F1 & P & R & F1 & N \\
        \hline\hline
        5 o Clock Shadow & 0.72 & 0.79 & 0.75 & 0.00 & 0.00 & 0.00 & 0.71 & 0.67 & 0.69 & 0.66 & 0.20 & 0.30 & 2345 \\
        Arched Eyebrows & 0.74 & 0.69 & 0.72 & 0.75 & 0.40 & 0.52 & 0.63 & 0.49 & 0.55 & 0.65 & 0.59 & 0.62 & 5134 \\
        Attractive & 0.79 & 0.86 & 0.83 & 0.87 & 0.47 & 0.61 & 0.77 & 0.81 & 0.79 & 0.78 & 0.79 & 0.79 & 10332 \\
        Bags Under Eyes & 0.67 & 0.52 & 0.59 & 0.60 & 0.07 & 0.12 & 0.54 & 0.47 & 0.50 & 0.62 & 0.29 & 0.40 & 4120 \\
        Bald & 0.74 & 0.48 & 0.58 & 0.00 & 0.00 & 0.00 & 0.73 & 0.46 & 0.56 & 0.74 & 0.23 & 0.35 & 410 \\
        Bangs & 0.84 & 0.86 & 0.85 & 0.83 & 0.05 & 0.10 & 0.81 & 0.86 & 0.84 & 0.84 & 0.85 & 0.85 & 2913 \\
        Big Lips & 0.62 & 0.22 & 0.33 & 0.37 & 0.43 & 0.40 & 0.54 & 0.21 & 0.31 & 0.59 & 0.18 & 0.28 & 3044 \\
        Big Nose & 0.79 & 0.44 & 0.56 & 0.61 & 0.49 & 0.54 & 0.63 & 0.52 & 0.57 & 0.65 & 0.48 & 0.56 & 4940 \\
        Black Hair & 0.78 & 0.75 & 0.76 & 0.65 & 0.13 & 0.21 & 0.76 & 0.72 & 0.74 & 0.68 & 0.80 & 0.74 & 4143 \\
        Blond Hair & 0.82 & 0.85 & 0.84 & 0.88 & 0.05 & 0.09 & 0.77 & 0.87 & 0.82 & 0.77 & 0.86 & 0.82 & 3054 \\
        Blurry & 0.72 & 0.45 & 0.55 & 0.77 & 0.01 & 0.02 & 0.62 & 0.38 & 0.47 & 0.65 & 0.35 & 0.45 & 929 \\
        Brown Hair & 0.68 & 0.64 & 0.66 & 1.00 & 0.00 & 0.00 & 0.70 & 0.56 & 0.62 & 0.74 & 0.42 & 0.53 & 4792 \\
        Bushy Eyebrows & 0.79 & 0.67 & 0.73 & 0.93 & 0.00 & 0.01 & 0.58 & 0.45 & 0.51 & 0.73 & 0.43 & 0.54 & 2830 \\
        Chubby & 0.68 & 0.48 & 0.57 & 0.62 & 0.04 & 0.07 & 0.50 & 0.50 & 0.50 & 0.63 & 0.29 & 0.40 & 1215 \\
        Double Chin & 0.70 & 0.50 & 0.59 & 0.57 & 0.01 & 0.02 & 0.51 & 0.50 & 0.50 & 0.69 & 0.29 & 0.40 & 975 \\
        Eyeglasses & 0.97 & 0.96 & 0.97 & 0.64 & 0.23 & 0.34 & 0.90 & 0.86 & 0.88 & 0.84 & 0.45 & 0.58 & 1380 \\
        Goatee & 0.81 & 0.69 & 0.75 & 0.60 & 0.01 & 0.01 & 0.79 & 0.61 & 0.69 & 0.67 & 0.17 & 0.28 & 1460 \\
        Gray Hair & 0.81 & 0.70 & 0.75 & 1.00 & 0.00 & 0.00 & 0.77 & 0.68 & 0.72 & 0.82 & 0.57 & 0.67 & 966 \\
        Heavy Makeup & 0.88 & 0.92 & 0.90 & 0.87 & 0.71 & 0.78 & 0.80 & 0.91 & 0.85 & 0.80 & 0.88 & 0.84 & 7751 \\
        High Cheekbones & 0.92 & 0.80 & 0.86 & 0.78 & 0.87 & 0.82 & 0.81 & 0.78 & 0.79  & 0.75 & 0.87 & 0.81 & 8926 \\
        Male & 0.97 & 0.98 & 0.98 & 0.91 & 0.85 & 0.88 & 0.94 & 0.95 & 0.94 & 0.94 & 0.93 & 0.93 & 8443 \\
        Mouth Slightly Open & 0.94 & 0.94 & 0.94 & 0.73 & 0.96 & 0.83 & 0.84 & 0.69 & 0.76 & 0.86 & 0.92 & 0.89 & 9569 \\
        Mustache & 0.72 & 0.49 & 0.59 & 0.00 & 0.00 & 0.00 & 0.52 & 0.34 & 0.41 & 0.42 & 0.04 & 0.08 & 1002 \\
        Narrow Eyes & 0.51 & 0.67 & 0.58 & 0.64 & 0.23 & 0.33 & 0.40 & 0.03 & 0.06 & 0.33 & 0.00 & 0.00 & 1491 \\
        No Beard & 0.97 & 0.98 & 0.98 & 0.87 & 0.97 & 0.92 & 0.95 & 0.98 & 0.96 & 0.91 & 0.95 & 0.93 & 16326 \\
        Oval Face & 0.67 & 0.29 & 0.40 & 0.87 & 0.00 & 0.01 & 0.58 & 0.31 & 0.41 & 0.51 & 0.39 & 0.44 & 5564 \\
        Pale Skin & 0.58 & 0.66 & 0.62 & 0.88 & 0.06 & 0.11 & 0.71 & 0.32 & 0.44 & 0.63 & 0.38 & 0.48 & 856 \\
        Pointy Nose & 0.65 & 0.45 & 0.53 & 0.74 & 0.02 & 0.04 & 0.55 & 0.34 & 0.42 & 0.61 & 0.24 & 0.35 & 5658 \\
        Receding Hairline & 0.64 & 0.43 & 0.52 & 0.60 & 0.01 & 0.02 & 0.64 & 0.41 & 0.50 & 0.54 & 0.43 & 0.48 & 1429 \\
        Rosy Cheeks & 0.77 & 0.40 & 0.52 & 1.00 & 0.00 & 0.00 & 0.51 & 0.58 & 0.54 & 0.54 & 0.48 & 0.50 & 1358 \\
        Sideburns & 0.84 & 0.75 & 0.79 & 0.00 & 0.00 & 0.00 & 0.84 & 0.65 & 0.73 & 0.75 & 0.20 & 0.32 & 1366 \\
        Smiling & 0.95 & 0.90 & 0.92 & 0.83 & 0.94 & 0.88 & 0.86 & 0.80 & 0.83 & 0.83 & 0.91 & 0.87 & 9601 \\
        Straight Hair & 0.60 & 0.41 & 0.49 & 0.00 & 0.00 & 0.00 & 0.55 & 0.40 & 0.47 & 0.52 & 0.26 & 0.34 & 4082 \\
        Wavy Hair & 0.68 & 0.64 & 0.66 & 0.57 & 0.22 & 0.32 & 0.66 & 0.62 & 0.64 & 0.67 & 0.56 & 0.61 & 5492 \\
        Wearing Earrings & 0.77 & 0.59 & 0.67 & 0.80 & 0.01 & 0.02 & 0.72 & 0.58 & 0.64 & 0.76 & 0.46 & 0.57 & 3789 \\
        Wearing Hat & 0.87 & 0.89 & 0.88 & 0.87 & 0.14 & 0.25 & 0.84 & 0.88 & 0.86 & 0.89 & 0.82 & 0.86 & 939 \\
        Wearing Lipstick & 0.88 & 0.96 & 0.92 & 0.87 & 0.82 & 0.85 & 0.83 & 0.95 & 0.89 & 0.83 & 0.94 & 0.89 & 8860 \\
        Wearing Necklace & 0.51 & 0.15 & 0.23 & 0.00 & 0.00 & 0.00 & 0.48 & 0.06 & 0.10 & 0.38 & 0.01 & 0.02 & 2396 \\
        Wearing Necktie & 0.60 & 0.29 & 0.39 & 0.00 & 0.00 & 0.00 & 0.54 & 0.29 & 0.38 & 0.57 & 0.09 & 0.16 & 1442 \\
        Young & 0.87 & 0.97 & 0.92 & 0.78 & 0.98 & 0.87 & 0.86 & 0.96 & 0.90 & 0.86 & 0.95 & 0.90 & 14821 \\
        \hline
        micro avg & 0.84 & 0.76 & 0.80 & 0.79 & 0.51 & 0.62 & 0.78 & 0.71 & 0.74 & 0.79 & 0.68 & 0.73 & 176143 \\
        macro avg & 0.76 & 0.65 & 0.69 & 0.63 & 0.25 & 0.28 & 0.69 & 0.59 & 0.62 & 0.69 & 0.50 & 0.55 & 176143 \\
        weighted avg & 0.82 & 0.76 & 0.78 & 0.73 & 0.51 & 0.52 & 0.76 & 0.71 & 0.72 & 0.75 & 0.68 & 0.69 & 176143 \\
        samples avg & 0.83 & 0.75 & 0.78 & 0.79 & 0.51 & 0.60 & 0.78 & 0.70 & 0.72 & 0.78 & 0.67 & 0.70 & 176143 \\
         \hline
    \end{tabular}
    \label{table:classification_report_face_attributes}
\end{table}

\begin{table}[!ht]
    \centering
    \begin{tabular}{lr}
    \toprule
    {} &  GANonymization \\
    \midrule
    Bald                &        1.000000 \\
    Gray Hair           &        1.000000 \\
    Double Chin         &        0.998494 \\
    Blurry              &        0.996370 \\
    Pale Skin           &        0.996337 \\
    Wearing Hat         &        0.993348 \\
    Wearing Necktie     &        0.992764 \\
    Mustache            &        0.992661 \\
    Chubby              &        0.984813 \\
    Goatee              &        0.973311 \\
    Wearing Necklace    &        0.972358 \\
    Eyeglasses          &        0.966012 \\
    Sideburns           &        0.949251 \\
    Big Nose            &        0.899965 \\
    Receding Hairline   &        0.877510 \\
    Bags Under Eyes     &        0.852971 \\
    Big Lips            &        0.780942 \\
    Wearing Earrings    &        0.768467 \\
    Black Hair          &        0.729177 \\
    Bushy Eyebrows      &        0.721409 \\
    5 o Clock Shadow    &        0.636142 \\
    Straight Hair       &        0.630562 \\
    Bangs               &        0.620606 \\
    Rosy Cheeks         &        0.615530 \\
    Blond Hair          &        0.615213 \\
    Pointy Nose         &        0.516256 \\
    Brown Hair          &        0.480853 \\
    Wavy Hair           &        0.410118 \\
    Narrow Eyes         &        0.400334 \\
    Male                &        0.276199 \\
    Arched Eyebrows     &        0.097596 \\
    Mouth Slightly Open &        0.089010 \\
    High Cheekbones     &        0.083279 \\
    Heavy Makeup        &        0.054131 \\
    Wearing Lipstick    &        0.048915 \\
    Smiling             &        0.046791 \\
    Oval Face           &        0.044784 \\
    No Beard            &        0.031004 \\
    Attractive          &        0.028488 \\
    Young               &        0.001595 \\
    \bottomrule
    \end{tabular}
    \caption{The table shows the percentage of removed traits over the total number of available samples for each trait in the validation set of the CelebA dataset.}
    \label{tab:celeba_classification}
\end{table}

\end{document}